\documentclass[runningheads]{llncs}

\usepackage[mobile]{eccv}

\usepackage{eccvabbrv}

\usepackage{graphicx}
\usepackage{booktabs}

\usepackage{amsmath}
\usepackage{algpseudocode}
\usepackage{array}
\usepackage{multirow}
\usepackage{dsfont}
\usepackage[font={small}]{caption}
\usepackage{wrapfig,lipsum,booktabs}
\usepackage{adjustbox}
\usepackage[ruled, lined, linesnumbered, commentsnumbered, longend]{algorithm2e}
\usepackage{listings}

\usepackage{ wasysym }

\usepackage{color}
\usepackage{xcolor}
\definecolor{citecolor}{HTML}{0071bc}
\usepackage{xspace}

\newcommand{\myparagraph}[1]{\vspace{2pt} \noindent \textbf{#1}}

\usepackage{makecell}
\usepackage{pifont}
\newcommand{\cmark}{{\color{green}\ding{52}}}
\newcommand{\xmark}{{\color{red}\ding{56}}}
\newcolumntype{P}[1]{>{\centering\arraybackslash}p{#1}} 
\newcolumntype{M}[1]{>{\centering\arraybackslash}m{#1}} 

\DeclareMathOperator*{\argmax}{arg\,max}

\usepackage[accsupp]{axessibility}  

\definecolor{mygreen}{rgb}{.372, .545, .067}

\usepackage[pagebackref,breaklinks,colorlinks,citecolor=eccvblue]{hyperref}

\usepackage{orcidlink}

\newlength\savewidth

\newcommand\markerlessfootnote[1]{
    \begingroup
    \renewcommand\thefootnote{}\footnote{#1}
    \addtocounter{footnote}{-1}
    \endgroup
}

\begin{document}

\title{Editable Image Elements for \\ Controllable Synthesis} 

\titlerunning{Editable Image Elements for Controllable Synthesis}

\author{Jiteng Mu\inst{1*} \and
Micha\"el Gharbi\inst{2} \and
Richard Zhang\inst{2} \and
Eli Shechtman\inst{2} \and
Nuno Vasconcelos\inst{1} \and
Xiaolong Wang\inst{1} \and
Taesung Park\inst{2}
}

\authorrunning{J.~Mu et al.}

\institute{University of California, San Diego, USA 
\email{\{jmu,nuno,xiw012\}@ucsd.edu}\\
\and
Adobe Research, USA
\email{\{mgharbi,rizhang,elishe,tapark\}@adobe.com}
\url{https://jitengmu.github.io/Editable_Image_Elements/}}

\maketitle

\begin{abstract}
    Diffusion models have made significant advances in text-guided synthesis tasks. However, editing user-provided images remains challenging, as the high dimensional noise input space of diffusion models is not naturally suited for image inversion or spatial editing. In this work, we propose an image representation that promotes spatial editing of input images using a diffusion model. Concretely, we learn to encode an input into ``image elements'' that can faithfully reconstruct an input image. These elements can be intuitively edited by a user, and are decoded by a diffusion model into realistic images. We show the effectiveness of our representation on various image editing tasks, such as object resizing, rearrangement, dragging, de-occlusion, removal, variation, and image composition.
\end{abstract}

\markerlessfootnote{* This work was done during an internship at Adobe Research.}


\section{Introduction}

\begin{figure}[t]
  \includegraphics[width=\textwidth]{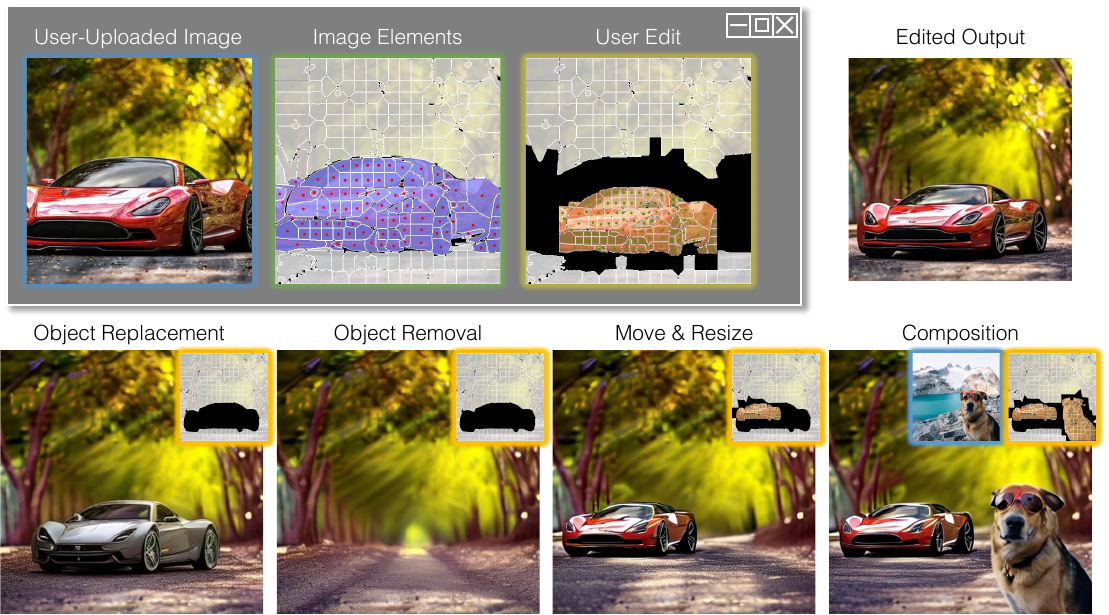}
    \vspace{-0.20in}
  \caption{We propose \textit{editable image elements}, a flexible representation that faithfully reconstructs an input image, while enabling various spatial editing operations. \textit{(top)} The user simply identifies interesting image elements (\textcolor{red}{red dots}) and edits their locations and sizes (\textcolor{mygreen}{green dots}). Our model automatically performs object-shrinking and de-occlusion in the output image to respect the edited elements. For example, the missing corners of the car are inpainted. \textit{(bottom)} More editing outputs are shown: object replacement, object removal, re-arrangement, and image composition.} 
    \vspace{-0.30in}
  \label{fig:teaser}
\end{figure}

High capacity diffusion models~\cite{ramesh2022hierarchical,saharia2022photorealistic,rombach2021highresolution} trained for the text-conditioned image synthesis task are reaching photorealism.
The strong image prior learned by these models is also effective for downstream image synthesis tasks, such as generating new scenes from spatial conditioning~\cite{meng2021sdedit,zhang2023adding}, or from a few example photos of a custom object~\cite{ruiz2023dreambooth,gal2023encoder,kumari2023multi}. 

However, while diffusion models are trained to generate images ``from scratch'', retrofitting them for image \textit{editing} 
remains surprisingly challenging.
One paradigm is to invert from image space into the noise space~\cite{song2020denoising,meng2021sdedit,parmar2023zero}. However, there is a natural tension between faithfully reconstructing the image and having an editable representation that follows the training distribution, leading to challenges in what types and how much noise and regularization to add. An alternative is to tune the diffusion model to condition on a representation of the image~\cite{zhang2023adding}, such as a ControlNet conditioned on edgemaps. However, while the diffusion model will adhere to the guidance, it may not capture properties of the input image that are absent in the guidance signal. Finally, one option is to tune the network on a set of images of a concept~\cite{ruiz2023dreambooth,gal2023encoder,kumari2023multi}. Although such methods generate new instances of the concept in novel situations, these are completely new images and not modifications of the original image.
Furthermore, in these existing workflows, the representations (the input noise map, or edge maps) are not amenable to precise spatial controls. Our goal is to explore a complementary representation to enable spatial editing of an input image.

To this end, we present an image editing framework that not only achieves faithful reconstruction of the input image at fast speed without optimization loops, but also allows for spatial editing of the input image. Our editing process begins by dividing each content of the input image into patch regions (Figure ~\ref{fig:teaser}) and encoding each patch separately. We represent the image as the collection of patch embeddings, sizes, and centroid locations, which are directly exposed to the user as controls for editing. The patch visualization provides intuitive control points for editing, as patches are movable, resizable, deletable, and most importantly delineated with semantically meaningful boundaries. The edited patch attributes are decoded into realistic images by a strong diffusion-based decoder. In particular, while our decoder is trained to preserve the input content as much as possible, it is still able to harmonize edited attributes into realistic images, even when some patch embeddings are missing or conflicting. Our method shares the same goal as image ``tokenization'' methods, such as VQGAN~\cite{esser2021taming} or the KL-autoencoder of Latent Diffusion Models~\cite{rombach2021highresolution}, as it aims to autoencode an input image into a collection of spatial embeddings that can faithfully reconstruct the original image. However, rather than adhering to the convolutional grid, our tokenization is spatially flexible, instead aligning to the semantically meaningful segments of the input image.

In summary, we propose a new image representation that supports
\begin{itemize}
    \item spatial editing of the input image content
    \item faithful reconstruction of the input image with low runtime
    \item various image editing operations like object removal, inpainting, resizing, and rearrangement
\end{itemize}
\section{Related Works}

\myparagraph{Image editing with diffusion models.} In image editing, as opposed to image synthesis, there exists extra challenges in preserving the content of the input images.
While text-to-image diffusion models are not designed to handle input images, several directions were developed to leverage the input content as inputs to the diffusion model, such as inverting the input image into the input noise map~\cite{mokady2023null,Wallace_2023_CVPR}, into the text embedding space~\cite{Kawar_2023_CVPR}, starting the denoising process from intermediate noise levels~\cite{meng2021sdedit,rombach2021highresolution} or conditioning on particular modalities like depthmap, pose estimation, or image embedding~\cite{zhang2023adding,ramesh2022hierarchical,gal2023encoder,voynov2022sketch}.
Also, there exist methods to handle specialized tasks in image editing, such as inpainting~\cite{Lugmayr_2022_CVPR,rombach2021highresolution,Yang_2023_CVPR}, text-guided modification~\cite{Brooks_2023_CVPR}, or customization from a few example images~\cite{ruiz2023dreambooth,kumari2023multi}.
However, most existing methods suffer from either inability to change the spatial layout~\cite{mokady2023null,Wallace_2023_CVPR,meng2021sdedit,Kawar_2023_CVPR} or loss of detailed contents~\cite{Brooks_2023_CVPR,gal2023encoder,ruiz2023dreambooth}.
Most notably, Self-Guidance~\cite{epstein2023selfguidance} proposes a way to enable spatial editing by caching attention maps and using them as guidance, but we observe that the edited outputs often fall off realism on many input images.

\myparagraph{Image Editing with Autoencoders.} The autoencoder design is a promising choice for image editing, where the input content is captured by the encoder and is assimilated with editing operations by the decoder to produce realistic images.
Since the encoding process is a feed-forward pass of the encoder, the speed is significantly faster than optimization-based methods~\cite{abdal2019image2stylegan,Kawar_2023_CVPR,ruiz2023dreambooth}, enabling interactive editing processes.
In GAN settings, Swapping Autoencoder~\cite{park2020swapping} showed structure-preserving texture editing by decomposing the latent space into structure and style code.
E4E~\cite{tov2021designing} designed an encoder for a pretrained StyleGAN~\cite{karras2020analyzing} generator.
In diffusion, Diffusion Autoencoder~\cite{preechakul2022diffusion} trains an encoder jointly with the diffusion-based decoder to capture the holistic content of the input image. 
However, there exists a fundamental trade-off between reconstruction accuracy and spatial editing capability, since the latent codes typically requires large spatial dimensions for good reconstruction, which cannot be easily edited with techniques like interpolation. 
Our method also falls under the autoencoder formulation, with an encoder processing individual image elements and a diffusion-based decoder mapping the edited image elements back into the pixel space. 
Still, we overcome the reconstruction-editing trade-off by representing the latent codes with continuous positional embeddings that are easily editable. 

\myparagraph{Layout Control in Generative Models} Controlling the generation process with layouts has been an active topic in deep image generative models~\cite{zhao2019image,li2021image,sun2021learning,sylvain2021object,jahn2021high,liu2022compositional,feng2022training,yang2022modeling,Li_2023_CVPR,avrahami2023spatext}, including segmentation-to-image synthesis models~\cite{isola2017image,wang2018high,park2019semantic}, or hierarchically refining the output image~\cite{ge2023expressive}. 
In particular, BlobGAN~\cite{epstein2022blobgan,wang2023blobgan} exposes objects knobs that can be continuously moved and resized, similar to our method. 
However, most existing methods cannot be used for editing input images, since the layout conditioning is insufficient to represent all input contents. 
Our method aims to enable image editing by comprehensively encoding all input contents into image elements in the form of superpixels~\cite{achanta2012slic}. 
\vspace{-0.15in}
\section{Method}

\begin{figure}[t]
    \centering
    \includegraphics[width=\linewidth]{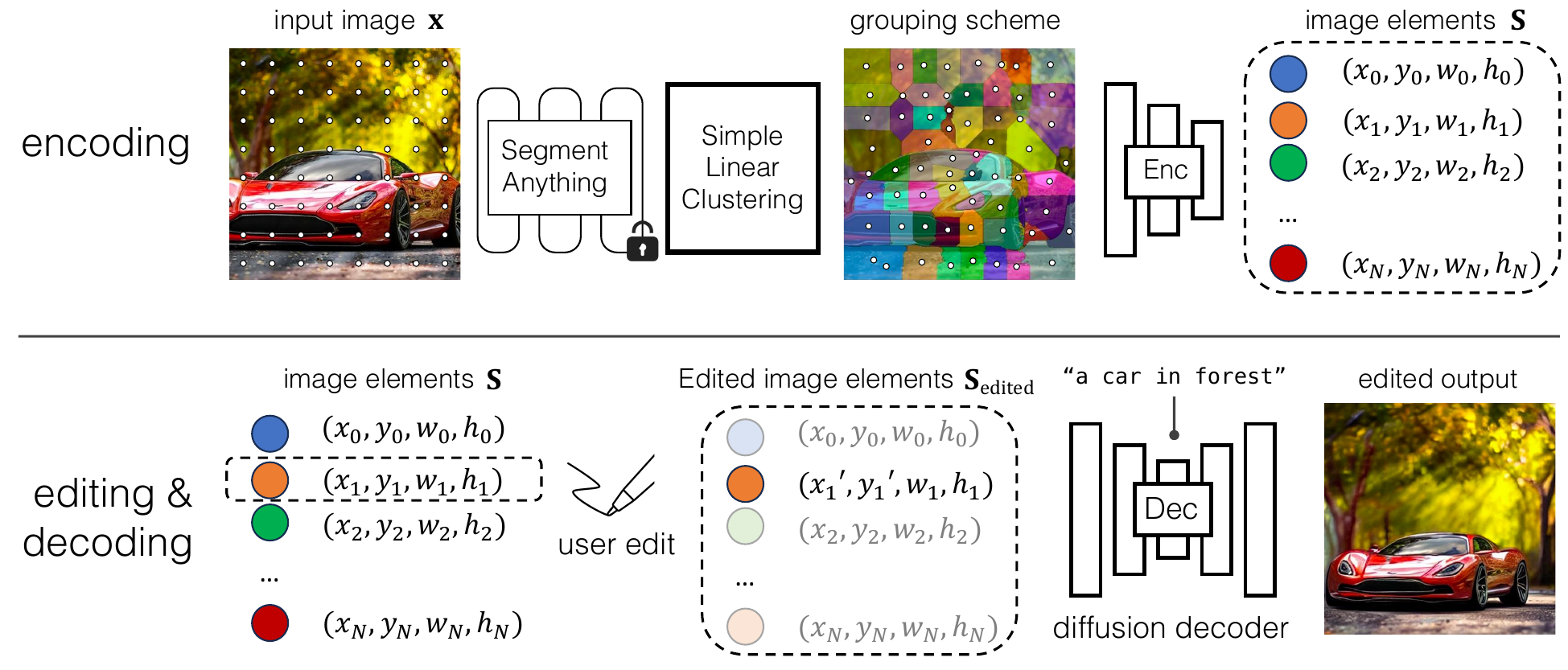}
    \vspace{-0.20in}
    \caption{ Overview of our image editing pipeline. \textit{(top)} To encode the image, we extract features from Segment Anything Model~\cite{kirillov2023segment} with equally spaced query points and perform simple clustering to obtain grouping of object parts with comparable sizes, resembling superpixels~\cite{achanta2012slic}. Each element is individually encoded with our convolutional encoder and is associated with its centroid and size parameters to form image elements. \textit{(bottom)} The user can directly modify the image elements, such as moving, resizing, or removing. We pass the modified image elements to our diffusion-based decoder along with a text description of the overall scene to synthesize a realistic image that respects the modified elements.}
    \label{fig: arch-stage1}
    \vspace{-0.3in}
\end{figure}

Our goal is to learn a representation that allows for photo-realistic image editing, particularly with spatial manipulations. To achieve this, the high-level idea is to divide an image into a set of manipulatable entities, such that each entity captures a part of an object or a stuff class, which can be intuitively edited for diverse operations like deletion, composition, or spatial re-arrangement. In addition, we train a decoder that conditions on the embeddings of these entities to produce realistic output RGB images. 

Our discussion will proceed in three sections. First, we present our strategy to define the image elements that can be easily edited by the user. Second, we design an encoder to capture the content of each image element independently by training an autoencoder that uses the image element as the bottleneck. Lastly, we replace the decoder part of the autoencoder with a more powerful text-guided diffusion model to create realistic images from the conditioning provided by the image elements. 

\subsection{Image Elements} \label{section-method: image elements}

We aim to represent an image $\mathbf{x}\in \mathds{R}^{H\times W\times 3}$ with ``image elements'' that capture the contents of the image while being amenable to editing at the same time. Specifically, each image element should correspond to an identifiable part of objects or ``stuff'' classes. Moreover, they should stay within the manifold of real image elements under editing operations such as deletion or rearrangement. For example, representing an image with a grid of latent codes, commonly used in latent autoencoder models~\cite{park2020swapping,esser2021taming,rombach2021highresolution} is not amenable for spatial editing, since the grid location of the unoccupied latent code cannot be left as blank before passed to the decoder. 

To this end, we perform grouping on the input image into disjoint and contiguous patches based on the semantic similarity and spatial proximity, represented as $\mathbf{A} = \{\mathbf{a}_1, \mathbf{a}_2, \cdots, \mathbf{a}_N\}$, where $\mathbf{a}_n\in \mathds{R}^{H_{n}\times W_{n}\times 3}$ is a cropped masked patch of the image.
For reference, we operate on images of size $H\times W=512\times 512$, using $N=256$ elements, with an average element size of $1024$ pixels.

\myparagraph{Obtaining image elements.} To divide the image into patches, we modify Simple Linear Iterative Clustering (SLIC)~\cite{achanta2012slic} to operate in the feature space of the state-of-the-art point-based Segmentation Anything Model (SAM)~\cite{kirillov2023segment}. We start with $N$ query points using $16\times 16$ regularly spaced points on the image. SAM can predict segmentation masks for each query point, making it a plausible tool for predicting image elements. However, the final predicted segmentation masks are not suitable as editable image elements, since the segments tend to vary too much in shape and size, and extreme deviation from the regular grid is not amenable to downstream encoding and decoding.
As such, we combine the predicted SAM affinity map $\mathbf{s}(m,n) \in [ 0, 1 ]$ with the Euclidean distance in spatial coordinates $\mathbf{d}(m,n)$, between pixel location index $m$ and query point $n$. Each pixel $m$ is grouped into a query element $n$

\begin{equation}
    g(m) = \argmax_{n \in \{{1, 2, \cdots, N}\}}[ \mathbf{s}(m, n)  - \beta \cdot \mathbf{d}(m, n)],
    \label{eq: slic algorithm}
\end{equation}

\noindent where hyperparameter $\beta$ is used to balance between feature similarity and spatial distance. The above formulation is equivalent to running one iteration of SLIC, thanks to the high-quality semantic affinity of the SAM feature space. According to $g(m)$, all pixels are then assigned to one of the $N$ query elements, resulting in a set of disjoint super-pixels $\mathbf{A}$. We post-process each patch $\textbf{a}_n$ by running connected components and selecting the largest region to ensure each patch is contiguous. In practice, this results in a small percentage of pixels $(\sim 0.1\%)$, corresponding to the smaller connected components, being dropped.

\myparagraph{Autoencoding image elements.} Now that we defined the grouping scheme for the image elements, we learn to encode the appearance of each patch by training an auto-encoder to reconstruct the image with the image element as the bottleneck. That is, we design encoder $\mathcal{E}$ and decoder $\mathcal{D}$ to incorporate the information from all image elements and reconstruct the image.

\begin{figure}[t]
    \centering
    \includegraphics[width=\linewidth]{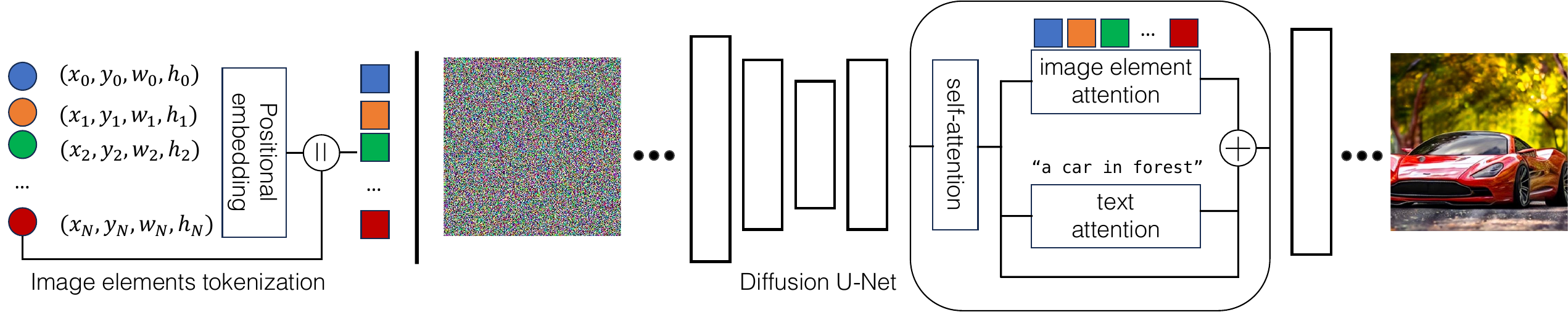}
    \vspace{-0.10in}
    \caption{Details of our diffusion-based decoder. First, we obtain positional embeddings of the location and size of the image elements, and concatenate them with the content embeddings to produce attention tokens to be passed to the diffusion model. Our diffusion model is a finetuned text-to-image Stable Diffusion UNet, with extra cross-attention layers on the image elements. The features from the text cross-attention layer and image element cross-attention layer are added equally to the self-attention features. Both conditionings are used to perform classifier-free guidance with equal weights.}
    \label{fig: arch-stage2}
    \vspace{-0.10in}
\end{figure}

\subsection{Content Encoder} \label{section-method: autoencoder with image element representatoin}

The goal of our encoding scheme is to disentangle the appearance and spatial information of the image elements, such that the elements can be later edited.
This is achieved by obtaining an appearance embedding on each patch \textit{separately}, to a convolutional encoder, such that the embedding is agnostic to its spatial location. To ensure the size parameters are decoupled from the feature, all patches $\mathbf{a}_n$ are resized to the same size before inputting to the encoder, and the patch embeddings are then obtained via the encoder. In addition, we collect patch properties $\textbf{p}_n$ --  centroid location $(x_n, y_n)$ and bounding box size $(w_n, h_n)$.

The convolutional encoder follows the architecture of KL Auto-encoder from Stable Diffusion, with 4 down-sampling layers.
The encoder is trained to maximize the informational content in its embedding by jointly training a lightweight decoder $\mathcal{D}_\text{light}$ to reconstruct the input image with the Euclidean loss.

\begin{equation}
\begin{aligned}
    \mathcal{E}^* = \arg &\min_\mathcal{E} \min_{\mathcal{D}_\text{light}} \ell_2 \big( \mathbf{x}, \mathcal{D}_\text{light}(\mathbf{S}) \big), \\
    & \text{where} \hspace{1mm} \mathbf{S} = \{(\mathcal{E}(\mathbf{a}_n), \mathbf{p}_n)\}, \text{encoded image elements}    
\end{aligned}
\end{equation}

In Section~\ref{section-exp: ablation}, we show through an ablation study that training such a lightweight transformer decoder is beneficial for learning better features.
The decoder is designed as a transformer, inspired by the Masked Auto-Encoder (MAE~\cite{he2022masked}) with 8 self-attention blocks. We insert 4 additional cross-attention layers in between to take the image elements as input. In addition, we input coordinate embeddings as the queries of the semantic decoder as a starting point.

\subsection{Diffusion Decoder} \label{section-method: diffusion decoder}
While the auto-encoder aforementioned produces meaningful image elements, the synthesized image quality is limited, since it is only trained with the MSE loss, resulting in blurry reconstructions. Furthermore, when the image elements are edited by the user $\mathbf{S}\rightarrow \mathbf{S}_\text{edited}$, we also empirically observe that realism is further compromised. 
To produce photo-realistic images from the image elements, the decoder should be powerful enough to fill in \textit{missing} information, unspecified by the edited image elements. We base such a decoder on the pretrained Stable Diffusion~\cite{rombach2021highresolution} model, modifying it to condition on our image elements.

\myparagraph{Stable Diffusion background.} Diffusion models are probabilistic models that define a Markov chain of diffusion steps to learn a data distribution by gradually denoising a normally distributed variable. As one of the most successful instantiation of diffusion models in the text2image generation models family, Stable Diffusion learns the distribution in the latent space to reduce the computation while preserving the image synthesis quality. Stable Diffusion is composed of two key components, 1) a variational auto-encoder to embed a clean image into latent space $\textbf{z}_0$, and 2) a base model $\mathcal{U}$ with parameters $\mathbf{\theta}_{\mathcal{U}}$ performs denoising of a noisy latent $\textbf{z}_{t}$, which is obtained by adding Gaussian noises over the clean latent $\textbf{z}_0$, over timestep $t \in \{0, 1, \cdots, T\}$, with $T=1000$,

Formally, the forward sampling process $q(\mathbf{z}_{t} | \mathbf{z}_{0})$ at time step $t$ is defined as,
\begin{equation}
    q(\mathbf{z}_{t} | \mathbf{z}_{0}) = \sqrt{\bar{\alpha}_{t}} \mathbf{z}_{0} + \sqrt{1 - \bar{\alpha}_{t}} \epsilon, \quad \epsilon \sim \mathcal{N}(\mathbf{0}, \mathbf{I}),
    \label{eq: forward sampling}
\end{equation}
where $\bar{\alpha}_t = \prod^{t}_{k} \alpha_k$ and $\alpha_1, \cdots, \alpha_t$ are pre-defined noise schedules. A sentence describing the picture is encoded to a text embedding $\textbf{C}$ using an additional text encoder and then fused into the base model via cross-attention. The training target of the base model $\mathcal{U}$ amounts to the following equation,
\begin{equation}
    \mathcal{L}_{SD}  = \mathbb{E}_{\mathbf{z}, \epsilon \sim \mathcal{N}(\mathbf{0}, \mathbf{I}), t} || \epsilon - \mathcal{U}(\textbf{z}_t, t, \textbf{C}; \mathbf{\theta}_{\mathcal{U}}) ||^2_2.
    \label{eq: SD train target}
\end{equation}

\noindent During inference, a clean latent $\textbf{z}_0$ can be generated by reversing the Gaussian process, going from pure Gaussian noise $\textbf{z}_T \sim \mathcal{N}(0, 1)$ to less noisy samples $\textbf{z}_{T-1}, \textbf{z}_{T-2}, \cdots, \textbf{z}_0 $ step-by-step, conditioned on the input text embedding $\textbf{C}$.

\myparagraph{Incorporating image elements.} The Stable Diffusion base model $\mathcal{U}$ is implemented with a UNet architecture, which contains a set of ResNet Blocks and Attention Blocks. The model takes a noisy latent $\mathbf{z}_t$ as input and predicts the corresponding noise $\hat{\epsilon}$. To incorporate the image element conditioning into the UNet model, we modify the attention blocks to jointly take the text embedding $\mathbf{C}$ and the image elements $\mathbf{S}$, as shown in Figure~\ref{fig: arch-stage2}.

Specifically, for each layer with a cross attention to text embeddings in the original UNet architecture, we insert a new cross attention layer with parameters $\theta_{\mathcal{S}}$ to take the image elements. Both cross attention layers, one on text and the other on image elements, are processed using the output features of the previous self-attention layers. The output features of the cross attention are then added to the self-attention layer features with equal weights. With this new layer, the original training target in Equation~\ref{eq: SD train target} now becomes, 
\begin{equation}
    \mathcal{L}_{SD}^\text{new} = \mathbb{E}_{\mathbf{z}, \epsilon \sim \mathcal{N}(\mathbf{0}, \mathbf{I}), t} || \epsilon - \mathcal{U}(\textbf{z}_t, t, \textbf{C}, \textbf{S}; \mathbf{\theta}_{\mathcal{U}}, \mathbf{\theta}_{\mathcal{S}}) ||^2_2.
    \label{eq: new SD train target}
\end{equation}

\noindent We initialize the set of parameters $\mathbf{\theta}_{\mathcal{U}}$ using the pre-trained model and the new set of introduced parameters $\mathbf{\theta}_{\mathcal{S}}$ are randomly initialized. Similar to the original Stable Diffusion where the text encoder is frozen, we freeze the parameters of both the content encoder and text encoder to ensure stability. We empirically justify the design decisions with ablation studies in Section~\ref{section-exp: ablation}.

We obtain an image $\hat{\mathbf{x}} =\mathcal{D}(\mathbf{z}_T, \mathbf{C},\mathbf{S})$, where $\mathcal{D}\triangleq \ocircle_{t=1}^T \mathcal{U}(\cdot, t, \cdot, \cdot)$ and $\mathbf{z}_T$ is randomly initialized noise.
We apply the same guidance weight of 3 on both types of cross-attentions. More details are included in the supplementary materials.

\myparagraph{Training with Image Element Dropout.} While the above training scheme is effective for reconstructing the original input from unedited image elements, $\hat{\mathbf{x}}=\mathcal{D}(\mathbf{S})$, we observe that the realism quickly degrades when generated an edited image $\hat{\mathbf{x}}_\text{edited}=\mathcal{D}(\mathbf{S}_\text{edited})$, as the edited elements can introduce distributional discrepancies unseen in training, such as overlapping, missing elements, or gaps between them. Therefore, we reduce the discrepancy between training and test time editing by designing a dropout scheme of image elements during training.

In practice, we employ Semantic SAM~\cite{li2023semantic}, an improved version of SAM enabling segment images at any desired granularity, to obtain a database of object masks. Then a random object mask is overlaid on the input image, and image elements overlapping with the mask are dropped out. 
However, due to the unwanted correlation between object edges and image elements boundaries, the model tends to inpaint object boundaries aligned with that of the dropped image elements, which is less preferred in many cases. 

To address this, we propose Random Partition, i.e., to divide image randomly to obtain the image elements. The insight behind is that the conditional probability of inpainting should ideally be independent of the image element partition. In practice, we simply obtain the random image elements partitioned from another sampled image.

\myparagraph{Supporting editing operations.} We show the following set of edits -- \textit{\textbf{delete}}, \textit{\textbf{move}}, and \textit{\textbf{resize}}. As the encoded appearance features $\mathcal{E}(\mathbf{a}_n)$ are decoupled from their spatial properties $\mathbf{p}_n$, these operations can be obtained by appropriate manipulations on each. So we can easily delete a subset of image elements or edit the positions and sizes $\mathbf{p}_n$ of the image elements. For deletion, instead of removing the image elements, we zero out both appearance features and their spatial information (after positional embedding) to maintain uniform input length during training. When performing moving or resizing, the image elements that collide with the edited image elements are automatically deleted. Intuitively, a pixel in the image space is always covered by at most one image element.

\begin{figure}
    \centering
    \includegraphics[width=\linewidth]{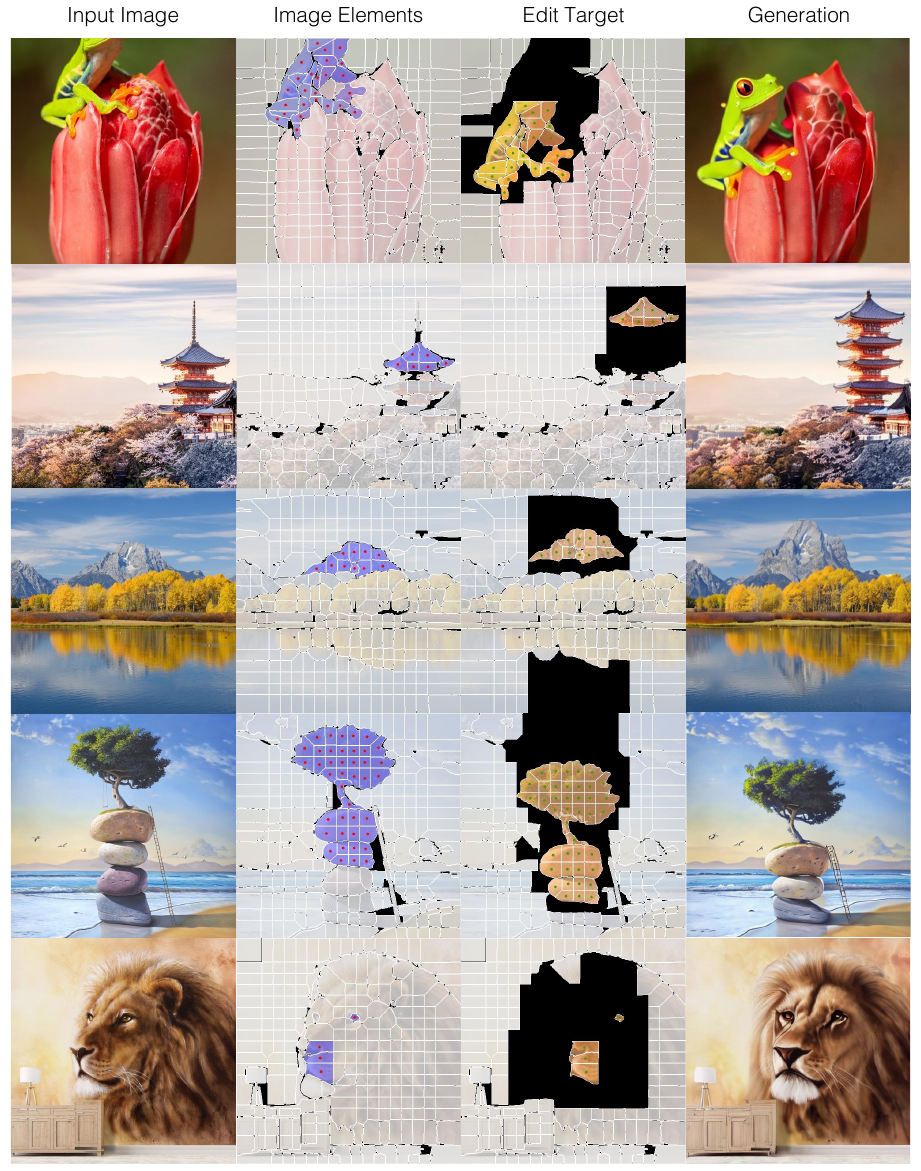}
    \vspace{-0.10in}
    \caption{The user can directly edit the image elements with simple selection, dragging, resizing, and deletion operations. The selected and edited elements are highlighted with \textcolor{red}{red} and \textcolor{mygreen}{green} dots at the centroid of each element.}
    \label{fig: exp figure 1}
    \vspace{-0.10in}
\end{figure}

\begin{figure}
    \centering
    \includegraphics[width=\linewidth]{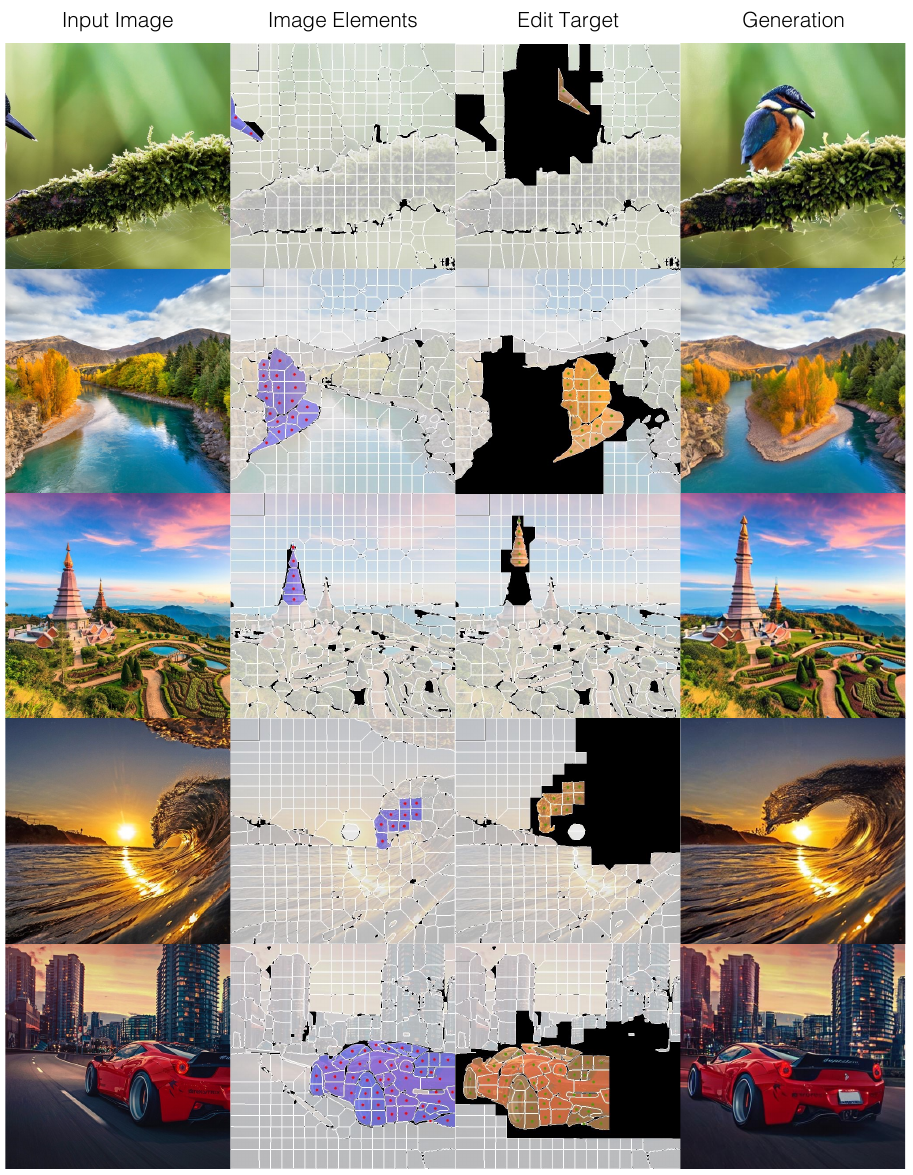}
    \vspace{-0.10in}
    \caption{The user can directly edit the image elements with simple selection, dragging, resizing, and deletion operations. The selected and edited elements are highlighted with \textcolor{red}{red} and \textcolor{mygreen}{green} dots at the centroid of each element.}
    \label{fig: exp figure 2}
    \vspace{-0.10in}
\end{figure}

\begin{figure}[t]
    \centering
    \includegraphics[width=\linewidth]{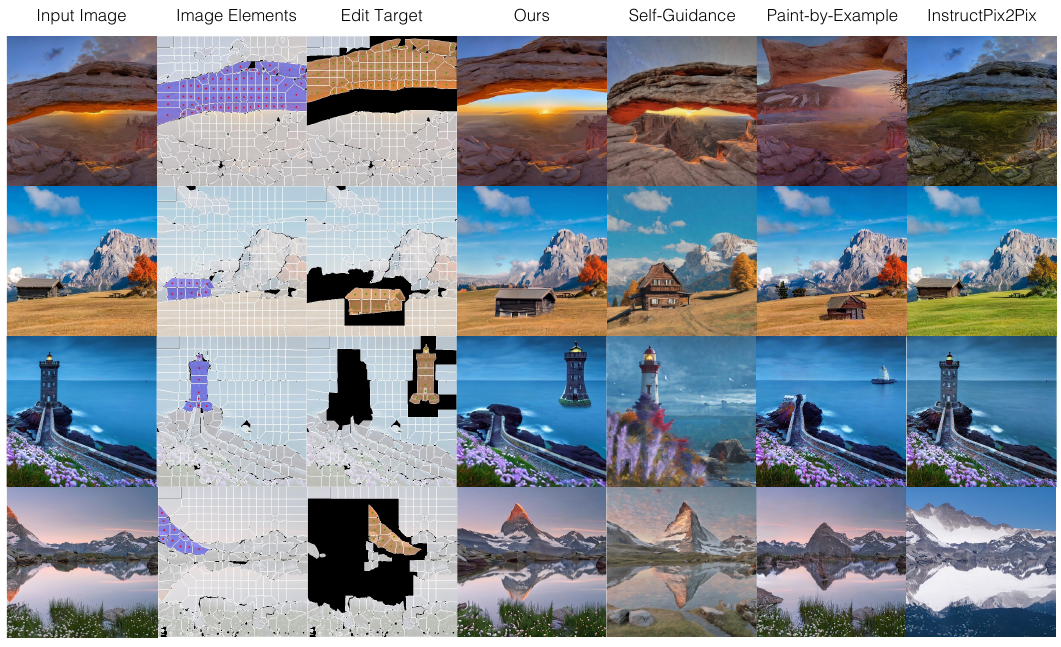}
    \vspace{-0.20in}
    \caption{Our method is compared to Self-guidance~\cite{epstein2023selfguidance}, Paint-by-Example~\cite{Yang_2023_CVPR}, and InstructPix2Pix~\cite{brooks2022instructpix2pix} on various edits. Our results attain superior results in preserving the details of the input as well as following the new edits. The baseline results show various types of failures, such as decline in image quality, floating textures, and unfaithful to the edits.}
    \label{fig: exp baseline}
    \vspace{-0.20in}
\end{figure}
\section{Experiments}

In this section, we show the rich examples on various image editing tasks, including object resizing, rearrangement, dragging, de-occlusion, object removal, and object variations. More details and results are presented in the appendix. 

\subsection{Dataset and Training Details} \label{section-exp: dataset}
Our dataset contains $3M$ images from the LAION Dataset~\cite{schuhmann2022laion}, filtered with an aesthetic score above $6.25$ and text-to-image alignment score above $0.25$. We randomly select $2.9M$ for training and evaluate the held-out $100$k data. The design of our content encoder follows the architecture of KL-autoencoder used in Stable Diffusion~\cite{rombach2021highresolution} and the transformer decoder follows Masked Autoencoders~\cite{he2022masked}. We elaborate more details in the supplementary materials.

For extracting image elements in Equation~\ref{eq: slic algorithm}, we empirically find $\beta=64$ yields a good balance between reconstruction quality and editability. For content encoder, we train the auto-encoder for 30 epochs with MSE loss. The decoder has eight self-attention layers with four additional cross attention layers taking the image elements as inputs. Our diffusion decoder is built on Stable Diffusion v1.5, trained with the same losses as Stable Diffusion. We report the results after around $180k$ iterations. We use classifier-free guidance with equal weights on text and image element conditioning to generate our examples; $\epsilon (z, \textbf{C}, \textbf{S}) = \epsilon (z, \emptyset, \emptyset) + w * [\epsilon (z, \textbf{C}, \textbf{S}) - \epsilon (z, \emptyset, \emptyset)]$. The examples in the paper are generated with $w = 3.0$ using $50$ sampling steps with the DDIM sampler.  

\begin{figure}[t]
    \centering
    \includegraphics[width=\linewidth]{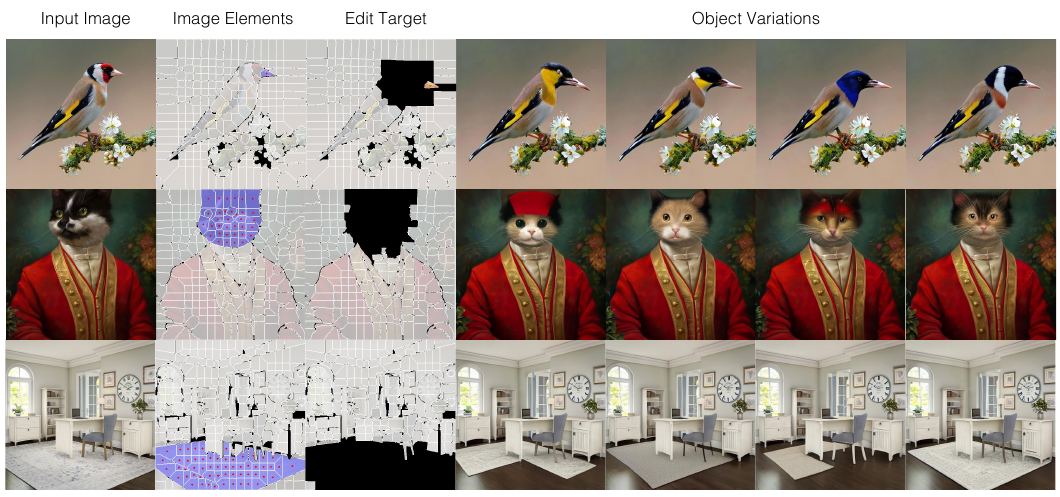}
    \vspace{-0.2in}
    \caption{Object variation. Our method supports object variation by deleting some image elements (shown in black in Target Edit), and performing inpainting guided by text prompt and the remaining image elements, such as the ``beak'' element in the top row.}
    \label{fig: exp object variation}
    \vspace{-0.10in}
\end{figure}

\begin{figure}[t]
    \centering
    \includegraphics[width=\linewidth]{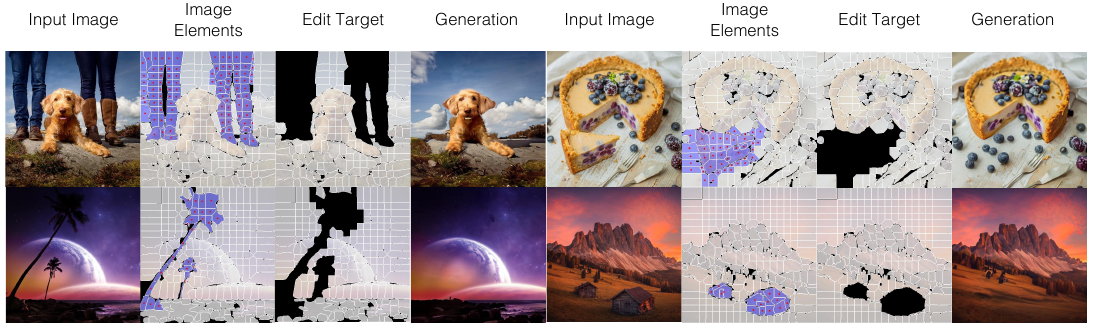}
    \vspace{-0.20in}
    \caption{Object removal. User selects elements to delete (shown in blue), and provide a text prompt pertaining to the background. Our diffusion decoder can generate content in the missing region (in black).}
    \label{fig: exp object removal}
    \vspace{-0.20in}
\end{figure}

\subsection{Spatial Editing} \label{section-exp: spatial edit}
To achieve spatial editing, the user directly modifies the image elements. Our diffusion-based decoder takes both image and text inputs to synthesize a realistic image that is faithful to the modified elements. As shown in Figure~\ref{fig: exp figure 1} and Figure~\ref{fig: exp figure 2}, we observe a number of interesting applications that are mostly unattempted by existing diffusion models.

We compared our method with various representative approaches, namely, gradient-based method Self-Guidance~\cite{epstein2023selfguidance}, exemplar-based inpainting method Paint-by-Example~\cite{Yang_2023_CVPR}, language instructed approach InstructPix2Pix~\cite{brooks2022instructpix2pix} in Figure~\ref{fig: exp baseline}. Self-Guidance extends prompt-to-prompt~\cite{hertz2022prompt} with additional gradient term to achieve spatial control for image editing. Though the method showcased interesting results on the stronger yet publicly unavailable Imagen~\cite{saharia2022photorealistic} model, we could only compare our result on the released SDXL~\cite{rombach2021highresolution} in Figure~\ref{fig: exp baseline}. We observe that with small guidance strength, the editing operation is not respected, and with higher guidance, the realism is affected. We hypothesize that the self-guidance requires very strong image prior similar to Imagen and is sensitive to the hyper parameters. In contrast, our editing results are more faithful to the input image and the editing operations. We also compare with Paint-by-Example, an  exemplar-guided image editing approach, by annotating images following the paper. To achieve spatial editing, we first crop a background region as the exemplar to inpaint the interesting area and then use the region of interest as the reference image to modify the target location. However, we empirically find the results usually lead to declined image quality for both stages. InstructPix2Pix is a language-based image editing method, which generates paired image editing examples using GPT3~\cite{brown2020language} and then trains a diffusion model for image editing. Though the method produces realistic images, it tends to either not responds to the edit or only modifies the global textures. In addition, it is also worth noting that using language only is naturally limited to achieve precise spatial editing.

\begin{wrapfigure}{r}{0.5\textwidth}
\centering
\vspace*{-1em}
\includegraphics[width=0.9\linewidth]{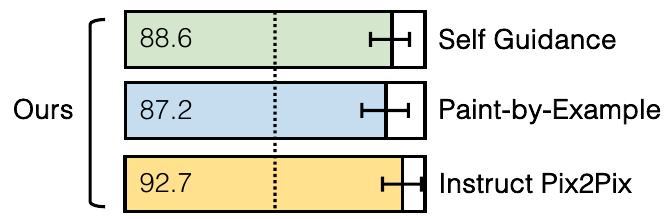}
\caption{Perceptual study where users are asked to choose which image better reflects both the editing and image quality.}
\label{fig: exp baseline user study}
\vspace*{-2em}
\end{wrapfigure}

Evaluating the quality of editing operation is a difficult task, as it is a combination of content preservation, adherence to task, and realism of the output images. Therefore, we run a Two-Alternative Forced Choice (2AFC) user study to quantify the quality of the editing results. We provide a pair of editing results with associated editing instructions, and ask the a user which image is better in terms of both image quality and faithfulness to the specified editing operation. As shown in Figure~\ref{fig: exp baseline user study}, our editing results are collected over 900 user judgments over three baselines, indicating that our method outperforms all baselines significantly. See the supplementary materials for the detailed experiment setting.


\subsection{Object Variations, Removal, and Composition} \label{section-exp: other edit}
Our method naturally supports object variation and removal by inpainting the specified region, shown in Figure~\ref{fig: exp object variation} and Figure~\ref{fig: exp object removal}. Since the model is conditioned on both the observed image elements and text, user can instruct model via text prompt to remove the object (inpaint with the background), or insert a new object in harmony with the context. Lastly, we can perform image composition with automatic harmonization by inserting elements from a different image, and removing the original elements in the overlapping region (Figure~\ref{fig:teaser}). 

\subsection{Ablation Studies} \label{section-exp: ablation}
\begin{figure}[t]
    \centering
    \includegraphics[width=\linewidth]{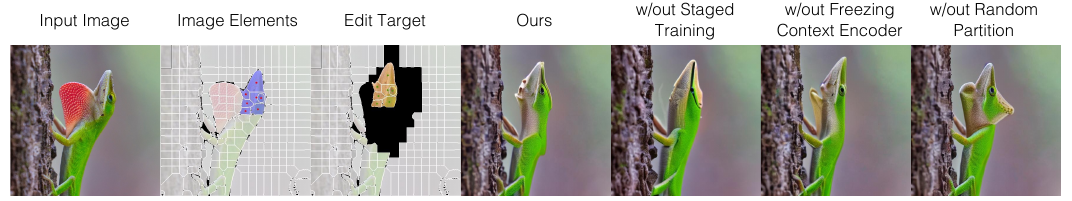}
    \vspace{-0.20in}
    \caption{Visualization of different design choices.}
    \label{fig: exp ablation}
    \vspace{-0.20in}
\end{figure}

In this section, we analyze various design choices as presented in Table~\ref{tab: ablation} and Figure~\ref{fig: exp ablation}. We compare both reconstruction and editing quality. For reconstruction, we provide each model with all image elements and run DDIM with 50 steps. For editing, similar to the user study conducted in Section~\ref{section-exp: spatial edit}, we present the user a pair of image editing results and ask the user which one is better regarding both faithfulness to the edit and image quality.  

Our proposed model is composed of two stages, with the first stage training the content encoder (Section~\ref{section-method: autoencoder with image element representatoin}) with a transformer decoder, and the second stage training the diffusion decoder while freezing the content encoder (Section~\ref{section-method: diffusion decoder}). The first question we study is: \emph{why is staged training necessary}? To show this, instead of training the content encoder with a transformer decoder first, we jointly train the content encoder and diffusion decoder together. We observe this variant is worse in both image reconstruction and editing quality. To further justify our design, we still do staged-training but do not freeze the content encoder when training the diffusion decoder. Though it shows better reconstruction compared to the joint-training, the overall quality is still inferior to our default setting. In terms of editing, the results are also less preferred compared to the default setting ($37.1\%$ compared to $62.9\%$).


Another trick for training the diffusion decoder is to overlay a partition to extract image elements. The reasoning for using random partitions, as opposed to the the actual partition of the image, is that we don't want the model to learn the correlation of image elements partition and image edges. Visually, we find the model trained without random partition tends to in-paint object boundaries aligned with that of the dropped image elements, which is less preferred than continuing the content outside the mask in many cases including object removal. As shown in Table~\ref{tab: ablation}, only $27.3\%$ of the editing results of the model without applying random partitions are more preferred.

\setlength{\tabcolsep}{4pt}
\begin{table}[!t]
\fontsize{7}{10pt}\selectfont
\begin{center}
\begin{tabular}{c c c | c c c c c c}
\hline\hline\noalign{\smallskip}
 \makecell{Staged \\ Training} & \makecell{Freeze \\ Content Encoder} & \makecell{Random \\ Partition} & MSE $\downarrow$ & PSNR $\uparrow$ & SSIM $\uparrow$ & LPIPS $\downarrow$ & FID $\downarrow$ & \makecell{ Preference \\ VS ours} $\uparrow$\\
\hline
\noalign{\smallskip}
\cmark & \cmark & \cmark & 0.0069 & 22.98 & 0.6354 & 0.3376 & 10.82 & -\\
\hline\noalign{\smallskip}
\xmark & \xmark & \cmark & 0.0138 & 19.74 & 0.5208 & 0.3697 & 11.91 & 34.2\% \\
\cmark & \xmark & \cmark & 0.0097 & 21.35 & 0.5962 & 0.3238 & 10.48 & 37.1\% \\
\cmark & \cmark & \xmark & 0.0066 & 23.15 & 0.6389 & 0.3262 & 9.75 & 27.3\% \\
\hline\hline
\end{tabular}
\end{center}
\vspace*{-0.8em}
\caption{Design differences. We study various design choices with both reconstruction and editing. ``Preference VS ours'' denotes the percentage of edited images that are preferred by MTurkers compared to our default setting. Our default setting achieves the best overall performance, both in image quality and in faithfulness to the editing.}
\label{tab: ablation}
\vspace*{-2em}
\end{table}
\setlength{\tabcolsep}{1.4pt}

\section{Discussion and Limitations}

We present the concept of editable image elements, which can be used to perform spatial editing on images using diffusion models. We show the inputs can be faithfully reconstructed by our modified diffusion model, and our approach enables various editing operations like moving, resizing, de-occlusion, removal, or variations. 


\textbf{Limitations} Since the reconstruction quality is not perfect, editing user-provided high resolution images remains challenging. Moreover, while our framework supports spatial editing, the appearance embeddings of the image elements are still not easily editable, such as rendering the image in different styles. Although not explored in the paper, image elements can be seen as a much more compact and controllable latent space than VQGAN. For example, a strong ``prior'' model could be trained to generate image elements, unifying the editing and synthesis models under the same framework. 

%
%
\bibliographystyle{splncs04}
\bibliography{main}

\clearpage
\section*{\Large{Appendices}}
\appendix

In this supplementary materials, we provide more details of the submission: We show additional editing results and pixel editing baselines in Section~\ref{supp section: additional editing comparison} complementing Section 4.2 in the paper; Furthermore, reconstruction evaluations are shown in Section~\ref{supp section: reconstruction comparison}; More implementation details, including architecture designs (paper Section 3.2 and Section 3.3), training recipes (paper Section 4.1), and image element partition algorithms (paper Section 3.1) are discussed in Section~\ref{supp section: implementation details}.

\section{Additional Editing Comparison} \label{supp section: additional editing comparison}
\begin{wrapfigure}{r}{0.5\textwidth}
\centering
\vspace*{-1em}
\includegraphics[width=0.9\linewidth]{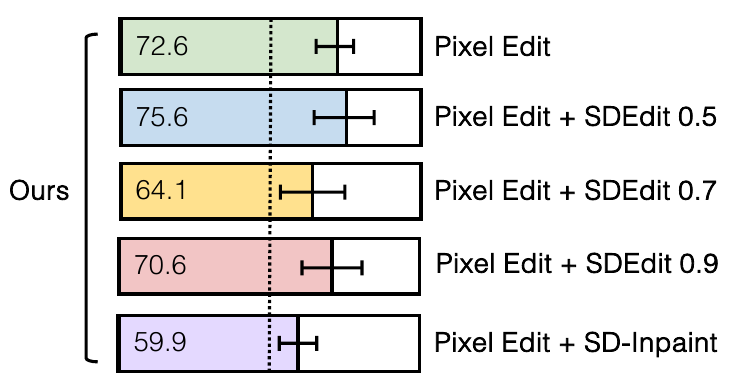}
\caption{Perceptual study where users are asked to choose which image better reflects both the editing and image quality. Results show that ours is preferred compared to all cases.}
\label{supp fig: exp user study}
\vspace*{-1em}
\end{wrapfigure}

In Figure 4, 5, and 6 of Section 4.2 in the main text, we have shown our editing results as well as comparisons to \textit{\textbf{Self-Guidance}}, \textit{\textbf{Paint-by-Example}}, and \textit{\textbf{InstructPix2Pix}}. Here we provide more details and show additional comparisons to these methods. Furthermore, we devise additional pixel editing related baselines built on our proposed image element partition, namely, \textit{\textbf{pixel editing}}, \textit{\textbf{pixel editing + SDEdit}}, and \textit{\textbf{pixel editing + SD-Inpaint}}. We run user studies on the pixel editing related baselines and show the results in Figure~\ref{supp fig: exp user study}. More visual comparisons are presented in Figure~\ref{supp fig: exp figure 1}, ~\ref{supp fig: exp figure 2}, ~\ref{supp fig: exp figure 3}, ~\ref{supp fig: exp figure 4}, ~\ref{supp fig: exp figure 5}, ~\ref{supp fig: exp figure 6}, ~\ref{supp fig: exp figure 7}, and ~\ref{supp fig: exp figure 8}. We elaborate on the detailed implementations of each baseline below.

\textbf{Self-Guidance}. The inversion of a real image is implemented following Self-Guidance, where a set of intermediate attention maps can be recorded by running a set of forward-process denoisings of a real input. Edits can then be performed with respect to the obtained attention maps. Since self-guidance modifies the gradient of each diffusion sampling step, it is sensitive to the hyperparameters choices. We observe that with small guidance strength, the editing operation is not respected, and with higher guidance, the realism is negatively affected. Instead of using separate parameters for individual images as in the paper, we employ the same parameter set for all testing images.

\textbf{Paint-by-Example}. Paint-by-Example requires a source image with a mask specifying the region to inpaint, plus a reference image for the target inpainting content. To achieve spatial editing, we take the pre-trained model and run inference as follows. We manually annotate each image with three regions: a source region for the object of interest, a target region for where to put the object, and a background region for inpainting the source region. In the first step, we treat the deliberately cropped background region as the reference to remove the object in the source region. Intuitively, this can be interpreted as object removal. Next, we directly regard the cropped source region as the reference image to inpaint the target location. Though the first step can potentially achieve object removal, it is difficult to find appropriate background region as reference for some images and edits, leading to degraded image realism. The second stage  poses further challenges in inpainting the source region content faithfully to the target location with expected size, especially when the source region is of low resolution or the source region contains only part of an object.

\textbf{InstructPix2Pix}. Though InstructPix2Pix is known to be great at texture transfer, we find its performance for spatial editing to be limited. We tried various prompts and found  the model tends to either not respond to the spatial instructions or only modifies the global textures. In addition, it is also worth noting that using language only is limiting to achieve precise spatial editing. In comparison, our proposed method directly takes coordinates to represent the locations and sizes, making it easy to change image elements following the requested spatial edits.

\textbf{Pixel editing}. Pixel editing is a simple baseline implemented by copy-pasting image elements in pixel space, where the image partition is the same as used in our algorithm. Specifically, we directly copy the image patches obtained with our algorithm to the target location and resize them as desired. As expected, pixel editing does not handle the source region properly, leading to object duplication and unrealistic images. In addition, it is also challenging to scale up the source region while maintaining high quality with simple pixel space resizing. To address the challenges, we further propose to use SDEdit and SD-Inpaint as described below. 

\textbf{Pixel editing + SDEdit}. SDEdit is a simple idea by first manipulating pixel space, and then adding Gaussian noise to the edited image and running the reverse sampling process for image synthesize. A sweet spot balancing realism and faithfulness can be identified for some specific intermediate time steps. Intuitively, adding more Gaussian noise and running the SDE for longer synthesizes more realistic images, but they are less faithful to the give input. To use SDEdit for spatial editing, we provide the masked pixel editing results as input, where the source region is left blank instead of maintaining the original pixels. We tested various time step choices (0.5, 0.7, 0.9) and find it is non-trivial to identify a single sweet spot for all images and editing operations. For a fair comparison, the diffusion model is chosen to be Stable Diffusion v1.5.

\textbf{Pixel editing + SD-Inpaiting}. We also test Stable Diffusion Inpainting model, where the UNet has 5 additional input channels (4 for the encoded masked-image and 1 for the mask itself) to inpaint the `holes' left in the source region.  We find though it produces reasonable inpainting results for some editing operations, it still suffers from scaling up the source region with high quality because the scaled source region is not modified by the diffusion model with masked input. In addition, we observe that it tends to keep inpainting another object rather than blending seamlessly with the background in the source region. Note that since we find SD-Inpainting model does not handle the small gaps between patches well, to get better results, we run the morphological operation on the masks to fill in the gaps first before passing it to the diffusion model.

\begin{figure}
\centering
\vspace*{-1em}
\includegraphics[width=\linewidth]{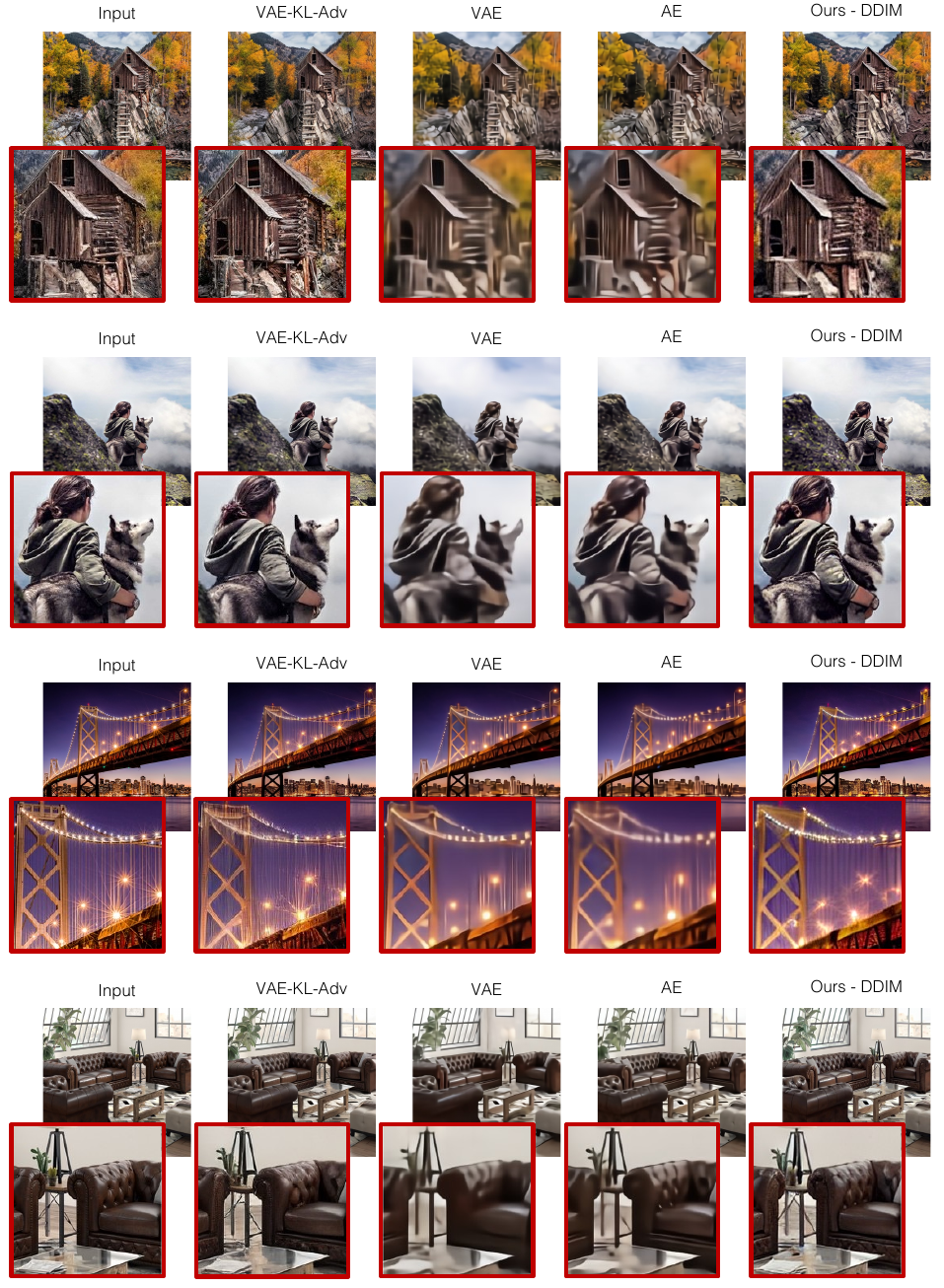}
\caption{Reconstruction comparisons. Ours-DDIM, trained with only L2 loss, preserves more details compared to autoencoder (AE) and variational autoencoder (VAE-KL), and achieves competitive results compared with variational autoencoder trained with adversarial loss (VAE-KL-Adv).}
\label{supp fig: exp reconstruction}
\vspace*{-2em}
\end{figure}

\setlength{\tabcolsep}{4pt}
\begin{table}[!t]
\fontsize{7}{10pt}\selectfont
\begin{center}
\begin{tabular}{c | c c | c c c c c }
\hline\hline\noalign{\smallskip}
 & Supervision & Bottleneck & MSE $\downarrow$ & PSNR $\uparrow$ & SSIM $\uparrow$ & LPIPS $\downarrow$ & FID $\downarrow$ \\
\hline
\noalign{\smallskip}
VAE-KL-Adv & L1+LPIPS+Adv & $16 \times 16 \times 32$ & 0.0052 & 24.56 & 0.6701 & 0.2180 & 5.96\\
\hline\noalign{\smallskip}
VAE-KL & L2 & $16 \times 16 \times 32$ & 0.0037 & 25.47 & 0.6754 & 0.4617 & 62.46 \\
AE & L2 & $16 \times 16 \times 32$ & 0.0027 & 27.55 & 0.7503 & 0.3759 & 55.65 \\
Ours - DDIM & L2 & $256 \times 32$ & 0.0069 & 22.98 & 0.6354 & 0.3376 & 10.82 \\
\hline\hline
\end{tabular}
\end{center}
\vspace*{-0.8em}
\caption{Reconstruction Comparison. Our method achieves better LPIPS and FID scores compared to AE and VAE (all trained with L2 loss), and competitive compared with variational autoencoder trained with adversarial loss. LPIPS and FID scores are more informative as we prefer faithful reconstruction instead of pixel-perfect results. Note MSE and PSNR are known to prefer blurry results as visualized in Figure~\ref{supp fig: exp reconstruction}. VAE-KL-Adv denotes variational autoencoder with KL divergence regularization plus losses following Latent Diffusion Model, VAE-KL for variational autoencoder with KL loss, AE for autoencoder, and ours-DDIM is with content encoder and diffusion decoder by running 50-step DDIM sampling steps. Numbers are computed on $5,000$ samples.}
\label{supp tab: reconstruction}
\vspace*{-2em}
\end{table}
\setlength{\tabcolsep}{1.4pt}

\section{Reconstruction Comparison} \label{supp section: reconstruction comparison}

To quantitatively evaluate the reconstruction of our method, we compare the reconstruction quality of the proposed method to various convolutional autoencoder approaches, as shown in Table~\ref{supp tab: reconstruction}. The bottleneck size of all methods is the same for a fair comparison. Qualitative comparisons are presented in Figure~\ref{supp fig: exp reconstruction}.

Specifically, our approach is composed of a content encoder and a diffusion decoder. The first stage trains the content encoder plus a lightweight transformer decoder. Then in the second stage, a diffusion decoder is learned with the content encoder frozen. The reconstruction results presented are obtained by providing all image patches to the content encoder to obtain corresponding image elements, then decoding with the diffusion decoder. We run 50 DDIM sampling steps for the results, showing that our method obtains better LIPIPS and FID compared to the autoencoder (AE) and variational autoencoder (VAE). From the visualizations in Figure~\ref{supp fig: exp reconstruction}, it is clear that our reconstruction maintains more details and is more visually appealing compared to AE and VAE, though showing slightly lower MSE and PSNR (which is widely known for preferring blurry results).  We also trained the VAE in an adversarial manner following the Latent Diffusion Model, showing on the top row for reference. Our method achieves competitive numbers, and from Figure~\ref{supp fig: exp reconstruction}, shows similar visual results (sharper details). Note that the simple L2 loss can be trained much faster than the adversarial loss. Adversarial training could potentially be used on our method for even better quality and we leave this for future work.

The input image size to all methods are $512 \times 512$. The architecture of all baselines are based on the autoencoder borrowed from the Latent Diffusion Model, which is a convolutional autoencoder with multiple residual blocks, downsampling layers and upsampling layers. The main differences are in that we use more downsampling and upsampling layers to obtain a compact latent representation ($32\times$ downsampling rather than $8\times$ downsampling). In addition, we also increase the number of channels of the bottleneck to 32 for a fair comparison. The VAE-KL-Adv is also included for reference, which is trained in an adversarial manner with an additional patch-based discriminator optimized to differentiate original images from reconstructions. To avoid arbitrarily scaled latent spaces, an Kullback-Leibler-term is implemented to regularize the latent. It also uses L1 loss combined with LPIPS loss for better reconstruction.

\section{Implementation Details} \label{supp section: implementation details}

In Section~\ref{supp section: architecture}, we provide more details of the content encoder and transformer decoder architectures, the pseudo-code for the fused attention block in our diffusion decoder. Detailed training recipes and hyper parameters are then presented in Section~\ref{supp section: training details}. We present the image partition algorithm and comparisons of different variants in Section~\ref{supp section: image element partition}.

\subsection{Architectures} \label{supp section: architecture}
\begin{figure}[t]
    \centering
    \includegraphics[width=\linewidth]{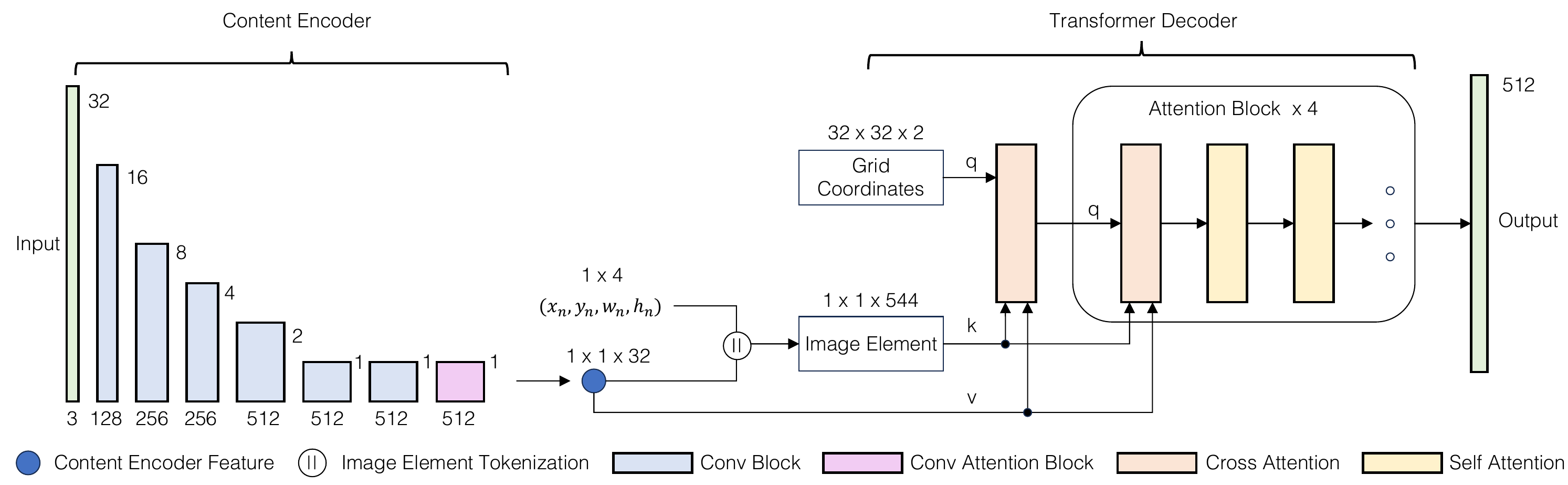}
    \vspace{-0.10in}
    \caption{Architecture for content encoder and transformer decoder. Only one image element is shown for simple illustration, but in practice, all image elements are jointly decoded.}
    \label{supp fig: exp autoencoder}
    \vspace{-0.10in}
\end{figure}

The architectures of the content encoder and transformer decoder described in paper Section 3.2 are illustrated in Figure~\ref{supp fig: exp autoencoder}. We also provide more details of the implementation of our fused attention block for the diffusion decoder (paper Section 3.3) in Algorithm~\ref{supp alg: diffusion decoder}.

\subsubsection{Content Encoder.}
Each image patch is first resized to $32 \times 32$ before input to the content encoder. This design ensures the features extracted are agnostic to positional and size information. The content encoder then maps the input image patch to a feature vector of 32 channels, as shown in Figure~\ref{supp fig: exp autoencoder}. The content encoder consists of multiple residual blocks, followed by convolutional downsampling layers with stride 2. One additional middle block consists of 2 Res-Blocks and 1 attention layer is implemented to further process the intermediate features. Then another convolutional layer outputs a feature of 32 channels for this patch. The overall architecture follows the convolutional encoder of the Latent Diffusion Model, but with less residual blocks at each level and a different output dimension. To compensate for the missing spatial information, the patch feature is combined with the a 4-dimensional vector, indicating its position and size, to form an image element.

\subsubsection{Transformer Decoder.}
To decode the image element, a set of grid coordinates of shape $32 \times 32 \times 2$ is passed to positional embedding to form the queries as the input of the transformer decoder. Then the image elements are served as keys and the features (excluding the spatial information) are used as values for the cross attention layer. Figure~\ref{supp fig: exp autoencoder} only shows one image element as an illustration. In practice, all image elements are jointly decoded using the transformer decoder. The network consists of 4 attention blocks, each is with 1 cross attention and 2 self-attention layers. All self-attention layers are with 512 channels and 16 heads, and cross attention layers with 512 channels and 1 head.

\begin{algorithm}[t]
\caption{Fused Attention Block}
\label{supp alg: diffusion decoder}
\definecolor{codeblue}{rgb}{0.25,0.5,0.5}
\lstset{
  backgroundcolor=\color{white},
  basicstyle=\fontsize{7.2pt}{7.2pt}\ttfamily\selectfont,
  columns=fullflexible,
  breaklines=true,
  captionpos=b,
  commentstyle=\fontsize{7.2pt}{7.2pt}\color{codeblue},
  keywordstyle=\fontsize{7.2pt}{7.2pt},
}
\begin{lstlisting}[language=python]
# x: input features
# context_image: image features mapped from image elements
# context_text: text features obtained from text encoders

# self attention
x = x + self_attention(norm_self(x))

# cross attention
if context_image is not None and context_text is not None:
    # fuse image and text attention outputs
    y_text = cross_attention_text(norm_text(x), context=context_text)
    y_image = cross_attention_image(norm_image(x), context=context_image)
    y = y_text + y_image
if context_image is not None and context_text is None:
    # image attention only
    y_image = cross_attention_image(norm_image(x), context=context_image)
    y = y_image
if context_text is not None and context_image is None:
    # text attention only
    y_text = cross_attention_text(norm_text(x), context=context_text)
    y = y_text
x = x + y

# feed forward
x = feedforward(norm_feedforward(x)) + x

\end{lstlisting}
\end{algorithm}

\subsubsection{Diffusion Decoder.}
While the auto-encoder aforementioned produces meaningful image elements, the synthesized image quality is limited. We modify pretrained Stable Diffusion model to condition on our image elements for better reconstruction and editing quality. The Stable Diffusion base model is implemented with a UNet architecture, which contains a set of residual blocks and attention blocks. To incorporate the image element into the UNet model, we modify the attention blocks to jointly take the text embedding and the image elements, as shown in Algorithm~\ref{supp alg: diffusion decoder}.

Specifically, for each layer with a cross attention to text embeddings in the original UNet architecture, we insert a new cross attention layer with parameters to take the image elements. Both cross attention layers, one on text and the other on image elements, are processed using the output features of the previous self-attention layers. The output features of the cross attention are then added to the self-attention layer features with equal weights.

\subsection{Training Details} \label{supp section: training details}

Our dataset contains $3M$ images from the LAION Dataset, filtered with an aesthetic score above $6.25$ and text-to-image alignment score above $0.25$. We randomly select $2.9M$ for training and leave the held-out $100$k data for evaluation. 

For the content encoder and lightweight transformer decoder, we train this autoencoder for $30$ epochs with MSE loss. We use the AdamW optimizer, setting the learning rate at $1e-4$, weight decay to $0.01$, and beta parameters to $(0.9, 0.95)$. The model is trained on $8$ GPUs, with a total batch size of $128$. In addition, all image elements are presented to the decoder for efficient training, which means no dropout is performed for this stage.

Our diffusion decoder is built on Stable Diffusion v1.5, trained with the same losses as Stable Diffusion. For the training phase, we use the AdamW
optimizer, setting the learning rate at $6.4e-5$, weight
decay to $0.01$, and beta parameters to $(0.9, 0.999)$. We report the results after around $180k$ iterations. The model is trained across $8$ GPUs, with a total batch size of 64. During inference, we use classifier-free guidance with equal weights on text $\textbf{C}$ and image element $\textbf{S}$ to generate our examples: $\epsilon (z, \textbf{C}, \textbf{S}) = \epsilon (z, \emptyset, \emptyset) + w * [\epsilon (z, \textbf{C}, \textbf{S}) - \epsilon (z, \emptyset, \emptyset)]$. The examples in the paper are generated with $w = 3.0$ using $50$ sampling steps with the DDIM sampler. During training, we randomly set only
$\textbf{C} = \emptyset$ for $30\%$ of examples, $\textbf{S} = \emptyset$ for $10\%$ of examples, and both $\textbf{C} = \emptyset$ and $\textbf{S} = \emptyset$ for $10\%$ of examples. 

\begin{algorithm}[t]
    \scriptsize
    \caption{\label{supp alg: image element partiation}Image Element Partition}
    \KwIn{Pretrained SAM model $M$,
    Image $\mathbf{x}$, Grid coordinates $Q$, Total number of iterations $T$, Centoid adjustment ratio $\beta_c$
    }
    \KwOut{Image partition $A = \{a_1, a_2, \cdots, a_N \}$, centoid locations $C = \{c_1, c_2, \cdots, c_N \}$, bounding box sizes $Z = \{z_1, z_2, \cdots, z_N \}$}
    
    $C_0 = Q$

    \For{$i = 1, 2, \cdots, T$}{

        $s \leftarrow M(\mathbf{x}, C_{i-1})$ \tcp{Compute SAM scores}

        $g \leftarrow \text{ComputeAssignment}(s)$ \tcp{SAM assignment}
        
        $C_i \leftarrow \text{ComputeCentroids}(g)$ \tcp{initial centroid}
        
        $\Tilde{C_i} \leftarrow \beta_c \cdot C_i + (1-\beta_c) \cdot Q$ \tcp{Centroid Adjustment}
        
        $\Tilde{g}  \leftarrow \text{ComputeAssignmentWithDistance}(s, \Tilde{C_i}, Q)$ \tcp{Distance regularization (Eq 1)}

        $\Tilde{g} \leftarrow \text{ConnectedComponent}(\Tilde{g} ) \text{ \textbf{if} $i=T-1$} $

        $C_i \leftarrow \text{ComputeCentroids}(\Tilde{g})$
    }
    \text{~}
    
    \tcp{Compute Outputs}
    $A \leftarrow \text{ComputePatches}(\Tilde{g})$
    
    $C \leftarrow \text{ComputeCentroids}(\Tilde{g})$
    
    $Z \leftarrow \text{ComputeSizes}(\Tilde{g})$
\end{algorithm}

\begin{figure}[t]
    \centering
    \includegraphics[width=\linewidth]{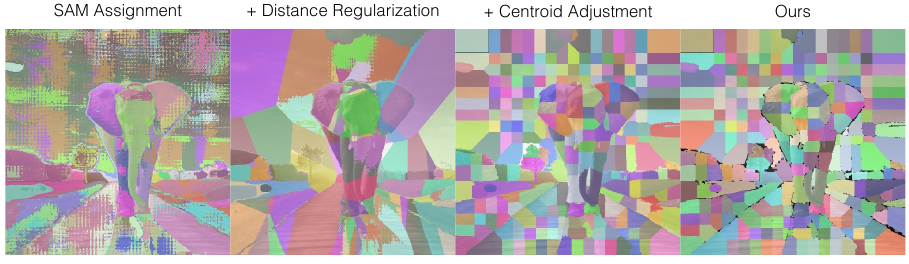}
    \vspace{-0.10in}
    \caption{Image element partition algorithm intermediate results. We show the step-by-step results complementing Algorithm~\ref{supp alg: image element partiation}. Starting from the noisy scores produced by SAM, distance regularization, centroid adjustment, morphological operation plus connected components are used sequentially to produce our final partitions.}
    \label{supp fig: exp partition steps}
    \vspace{-0.10in}
\end{figure}

\subsection{Image Element Partition} \label{supp section: image element partition}

We here provide more details of the image partition algorithm for Section 3.1 in the paper. Specifically, we implement Algorithm~\ref{supp alg: image element partiation} for extracting image elements, with intermediate outputs shown in Figure~\ref{supp fig: exp partition steps}.

To divide the image into patches, we borrow the insight of the Simple Linear Iterative Clustering (SLIC) to operate in the feature space of the state-of-the-art point-based Segmentation Anything Model (SAM). We start with $N=256$ query points using $16\times 16$ regularly spaced points $Q$ on the image, resulting in at most $256$ image element partitions in the end. To start with, SAM takes an image $\textbf{x}$ and initial centroids $C_0$ as inputs, and predicts association scores $s$ for each query point and pixel locations. Then it follows by computing the cluster assignment $g$ for all pixel locations. However, since the segments tend to vary too much in shape and size, and extreme deviation from the regular grid is not amenable to downstream encoding and decoding, see Figure~\ref{supp fig: exp partition steps} SAM Assignment. So we propose to add distance regularization as described in Eq 1 in the paper with hyper parameters $\beta$ balancing the distance and scores, visualized in Figure~\ref{supp fig: exp partition steps} Distance Regularization. Nevertheless, it is observed that centroids tend to collapse for semantically close regions. To address this, we further modify the obtained centroids by computing a linear interpolation of the centroids $C_i$ and grid coordiantes $Q$, with a hyper parameter $\beta_c=0.2$. This effectively avoids centroid collapse as illustrated in Figure~\ref{supp fig: exp partition steps} Centroid Adjustment. Finally, morphological operations and connected components are applied to remove small regions. We set the number of iterations to be $1$, but potentially more iterations could be used for better partitions.

\begin{figure}[h]
    \centering
    \includegraphics[width=\linewidth]{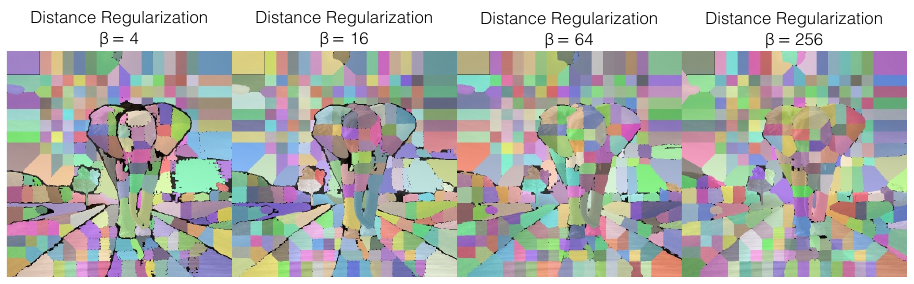}
    \vspace{-0.10in}
    \caption{Comparisons on different distance regularization strengths corresponding to Eq 1 in the paper. Smaller $\beta$ tends to drop more pixels and larger $\beta$ produces less accurate superpixels. $\beta=64$, achieving a sweet spot between reconstruction and editability, is used throughout the paper. }
    \label{supp fig: exp partition distance}
    \vspace{-0.10in}
\end{figure}

\begin{figure}[!bht]
    \centering
    \includegraphics[width=\linewidth]{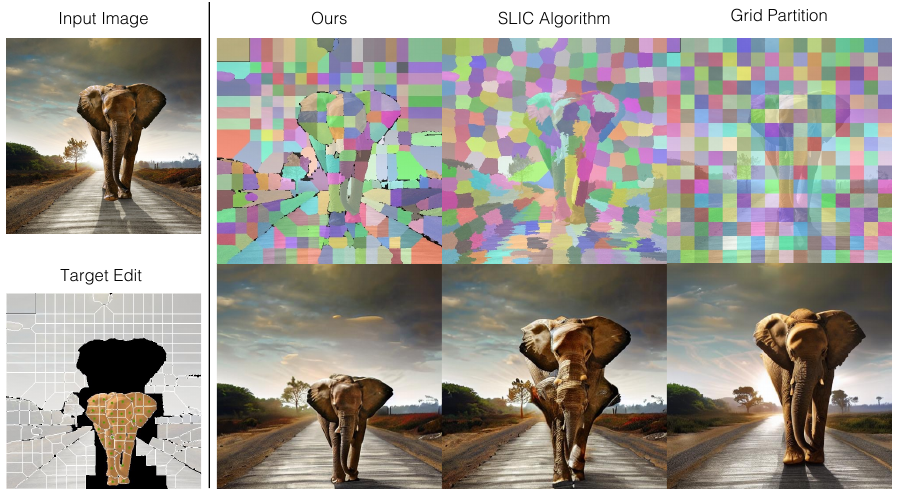}
    \vspace{-0.10in}
    \caption{Comparisons on different image partition algorithms. We test various image partitions on the same edit and results are shown on the right: ours, SLIC algorithm, and grid partition. Ours follows the edit operations and preserves object details correctly. In comparison, SLIC algorithm produces unrealistic images and grid partition fails to follow the edit.}
    \label{supp fig: exp partition comparison}
    \vspace{-0.10in}
\end{figure}

We also study various parameter choices for $\beta$ in Eq 1 as shown in Figure~\ref{supp fig: exp partition distance}. We empirically choose $\beta=64$, which achieves a good balance between reconstruction quality and editability. We also compare with various image partition algorithms, such as SLIC algorithm in pixel space and grid partition. From Figure~\ref{supp fig: exp partition comparison}, our algorithm yields best editing results, whereas other methods presents various types of failures. We observe that for SLIC algorithm, the image partitions tend to show wiggling boundaries and the superpixels obtained are also not well aligned with object boundaries as ours, making the learning harder. For grid partition, since the location and size parameters are constants across all images, it fails to relocate and resize objects. This again shows that our design of image element partition and resizing are the keys to learn a disentangled representation.

\begin{figure}[t]
    \centering
    \includegraphics[width=\linewidth]{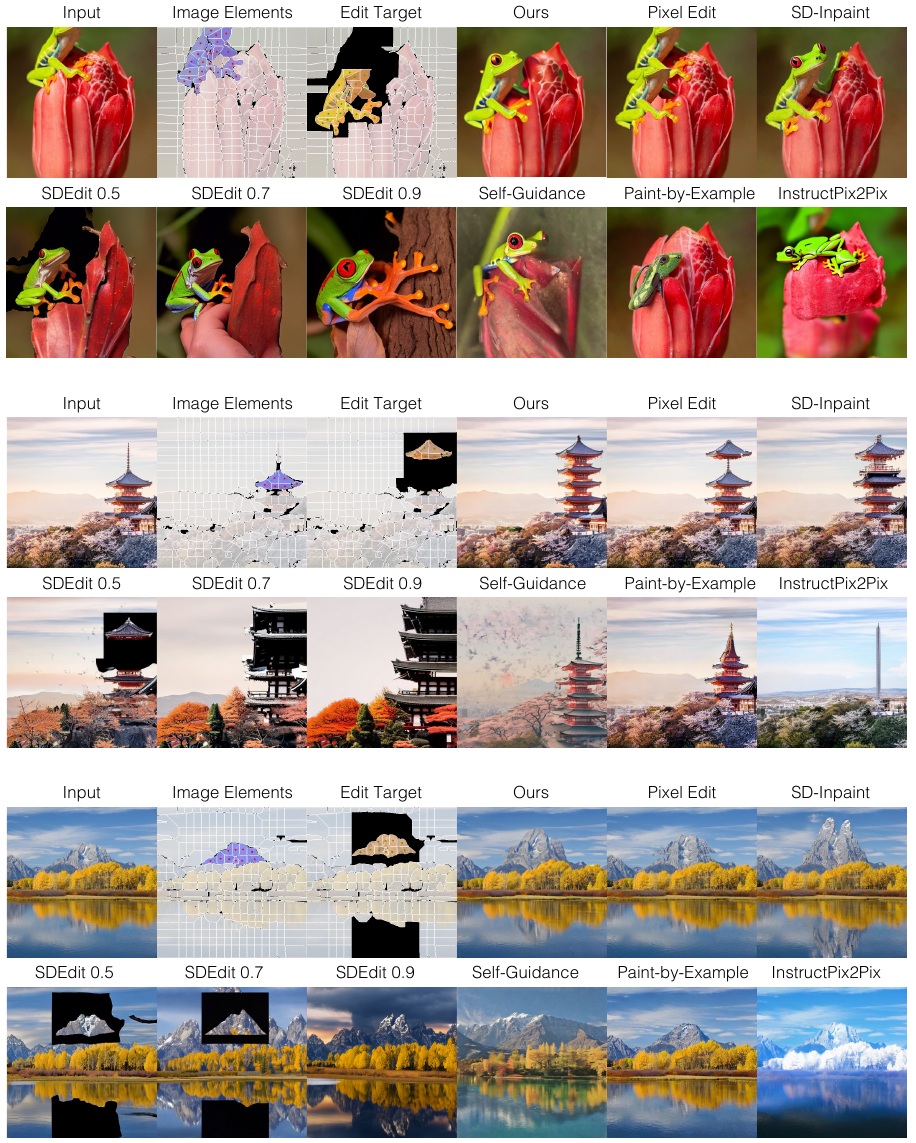}
    \vspace{-0.10in}
    \caption{Comparisons complementing Figure 4, 5, and 6. Our method is compared to pixel editing, pixel editing with Stable Diffusion Inpainting model (SD-Inpaint), pixel editing with SDEdit of various schedules, Self-guidance, Paint-by-Example, and InstructPix2Pix on various edits. Our results attain superior results in preserving the details of the input as well as following the new edits. The baseline results show various types of failures, such as decline in image quality, floating textures, and unfaithful to the edits.}
    \label{supp fig: exp figure 1}
    \vspace{-0.10in}
\end{figure}

\begin{figure}[t]
    \centering
    \includegraphics[width=\linewidth]{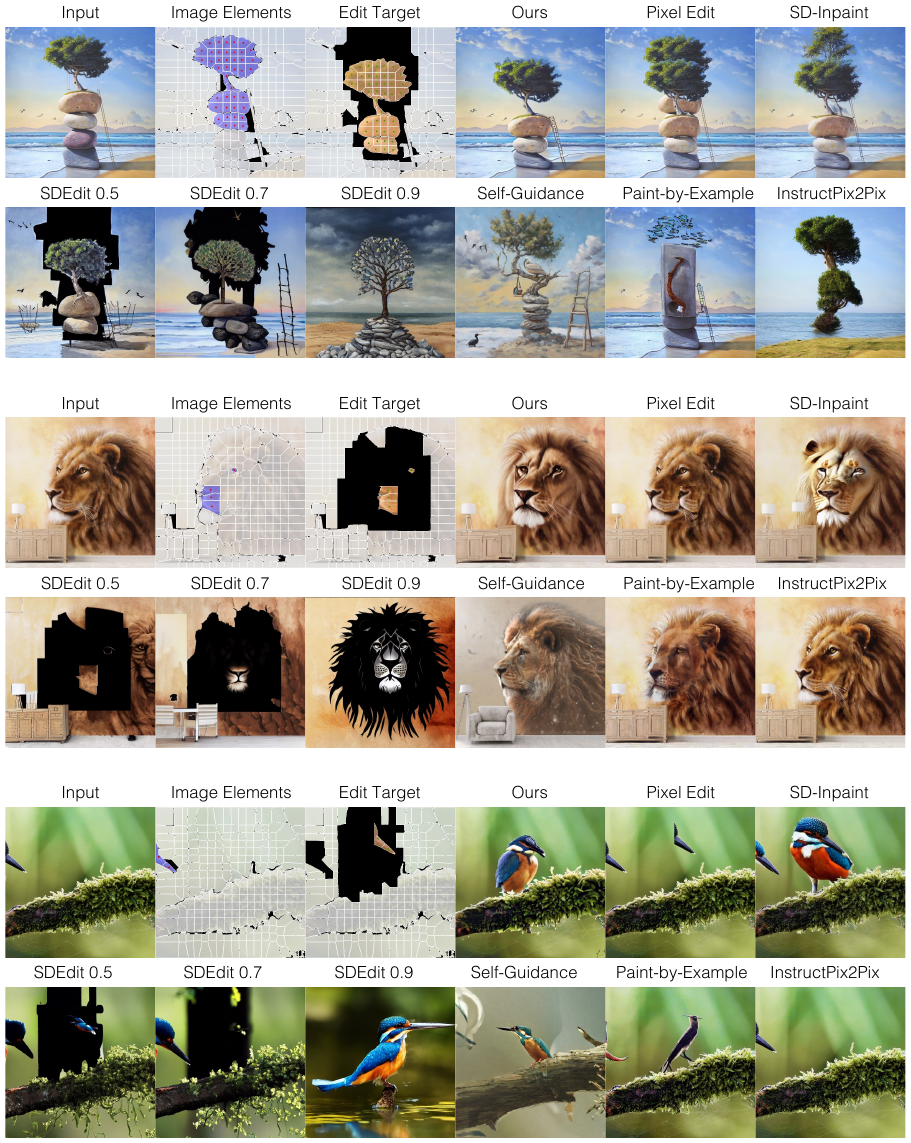}
    \vspace{-0.10in}
    \caption{Comparisons complementing Figure 4, 5, and 6. Our method is compared to pixel editing, pixel editing with Stable Diffusion Inpainting model (SD-Inpaint), pixel editing with SDEdit of various schedules, Self-guidance, Paint-by-Example, and InstructPix2Pix on various edits. Our results attain superior results in preserving the details of the input as well as following the new edits. The baseline results show various types of failures, such as decline in image quality, floating textures, and unfaithful to the edits.}
    \label{supp fig: exp figure 2}
    \vspace{-0.10in}
\end{figure}

\begin{figure}[t]
    \centering
    \includegraphics[width=\linewidth]{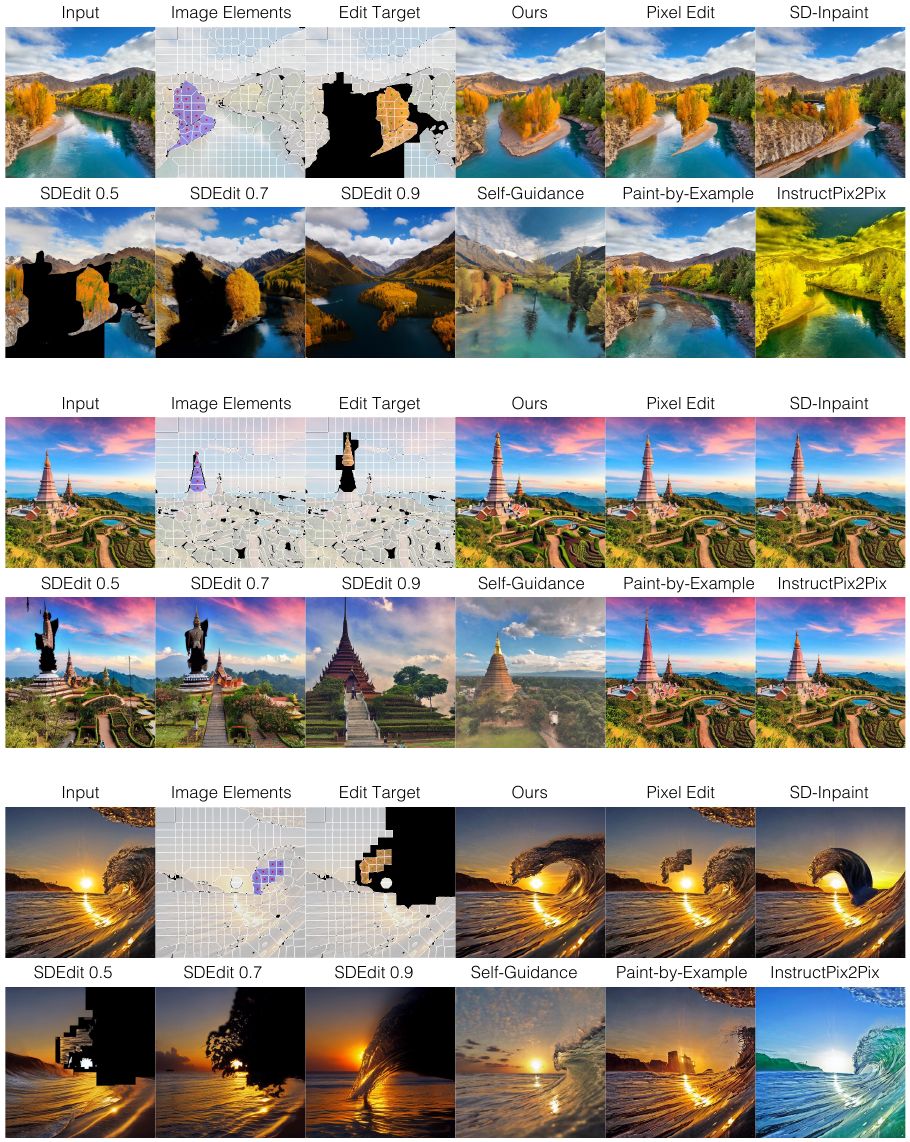}
    \vspace{-0.10in}
    \caption{Comparisons complementing Figure 4, 5, and 6. Our method is compared to pixel editing, pixel editing with Stable Diffusion Inpainting model (SD-Inpaint), pixel editing with SDEdit of various schedules, Self-guidance, Paint-by-Example, and InstructPix2Pix on various edits. Our results attain superior results in preserving the details of the input as well as following the new edits. The baseline results show various types of failures, such as decline in image quality, floating textures, and unfaithful to the edits.}
    \label{supp fig: exp figure 3}
    \vspace{-0.10in}
\end{figure}

\begin{figure}[t]
    \centering
    \includegraphics[width=\linewidth]{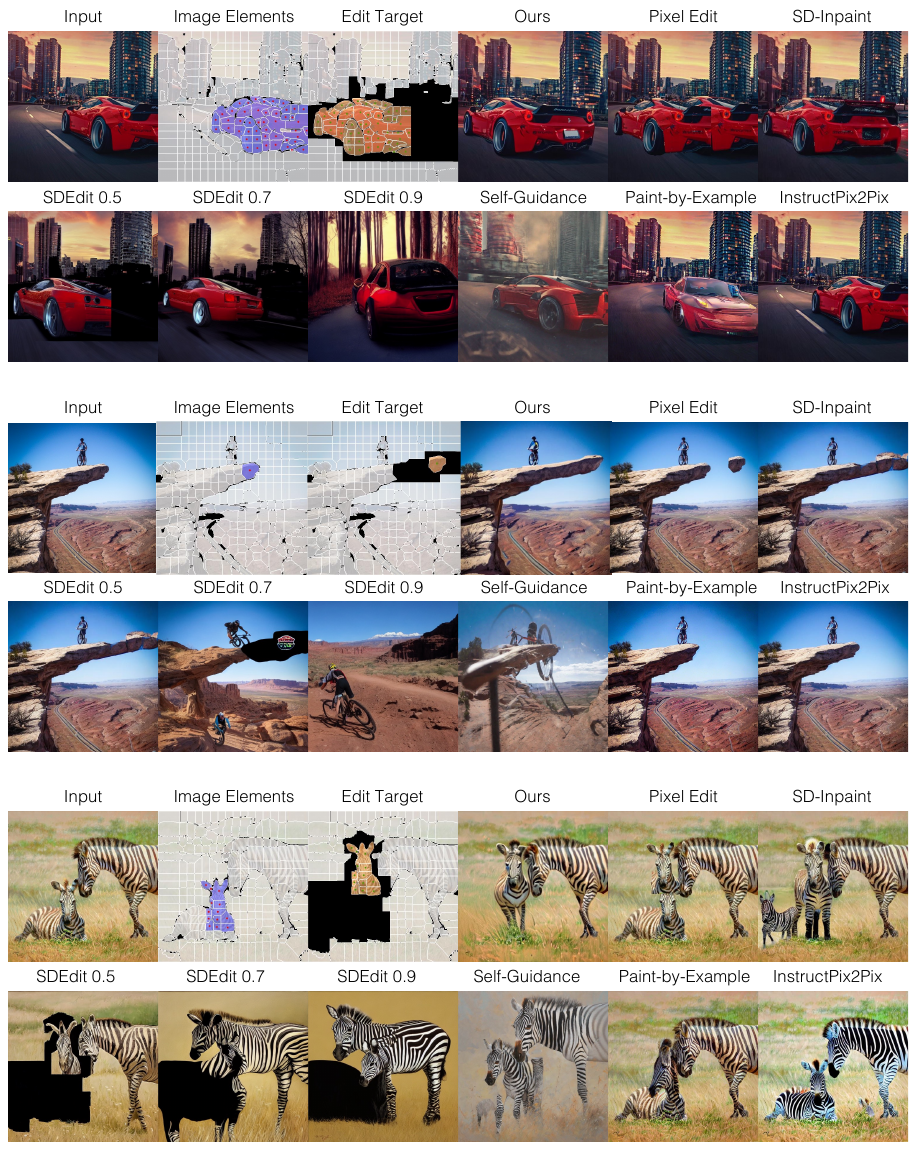}
    \vspace{-0.10in}
    \caption{Comparisons complementing Figure 4, 5, and 6. Our method is compared to pixel editing, pixel editing with Stable Diffusion Inpainting model (SD-Inpaint), pixel editing with SDEdit of various schedules, Self-guidance, Paint-by-Example, and InstructPix2Pix on various edits. Our results attain superior results in preserving the details of the input as well as following the new edits. The baseline results show various types of failures, such as decline in image quality, floating textures, and unfaithful to the edits.}
    \label{supp fig: exp figure 4}
    \vspace{-0.10in}
\end{figure}

\begin{figure}[t]
    \centering
    \includegraphics[width=\linewidth]{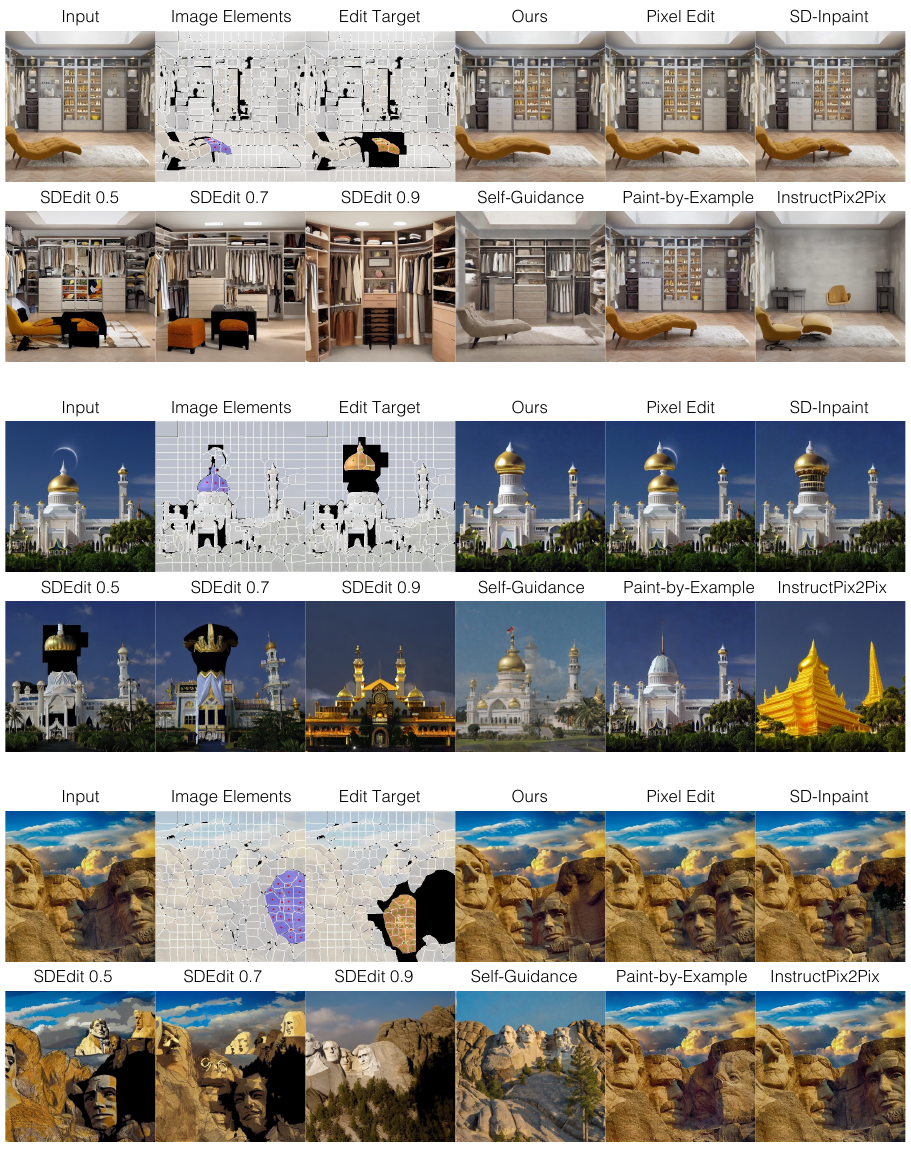}
    \vspace{-0.10in}
    \caption{Comparisons complementing Figure 4, 5, and 6. Our method is compared to pixel editing, pixel editing with Stable Diffusion Inpainting model (SD-Inpaint), pixel editing with SDEdit of various schedules, Self-guidance, Paint-by-Example, and InstructPix2Pix on various edits. Our results attain superior results in preserving the details of the input as well as following the new edits. The baseline results show various types of failures, such as decline in image quality, floating textures, and unfaithful to the edits.}
    \label{supp fig: exp figure 5}
    \vspace{-0.10in}
\end{figure}

\begin{figure}[t]
    \centering
    \includegraphics[width=\linewidth]{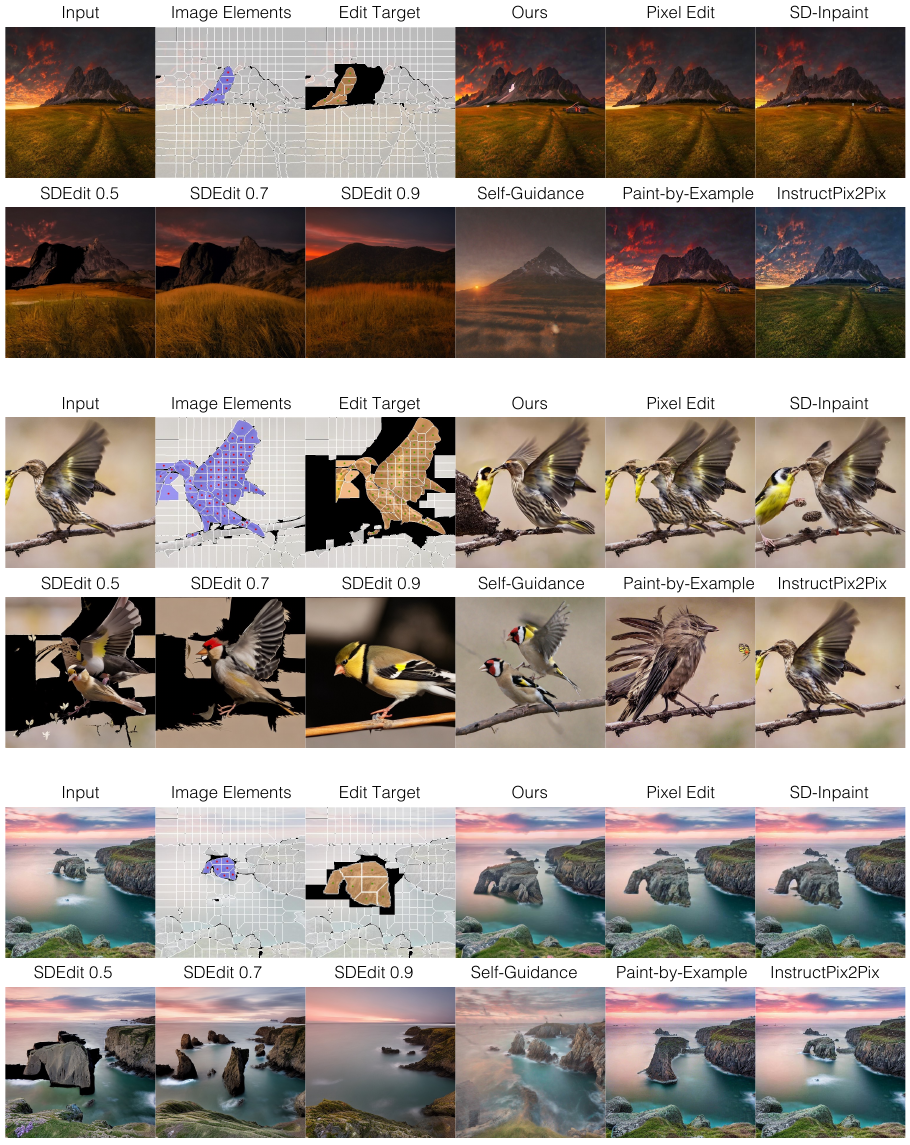}
    \vspace{-0.10in}
    \caption{Comparisons complementing Figure 4, 5, and 6. Our method is compared to pixel editing, pixel editing with Stable Diffusion Inpainting model (SD-Inpaint), pixel editing with SDEdit of various schedules, Self-guidance, Paint-by-Example, and InstructPix2Pix on various edits. Our results attain superior results in preserving the details of the input as well as following the new edits. The baseline results show various types of failures, such as decline in image quality, floating textures, and unfaithful to the edits.}
    \label{supp fig: exp figure 6}
    \vspace{-0.10in}
\end{figure}

\begin{figure}[t]
    \centering
    \includegraphics[width=\linewidth]{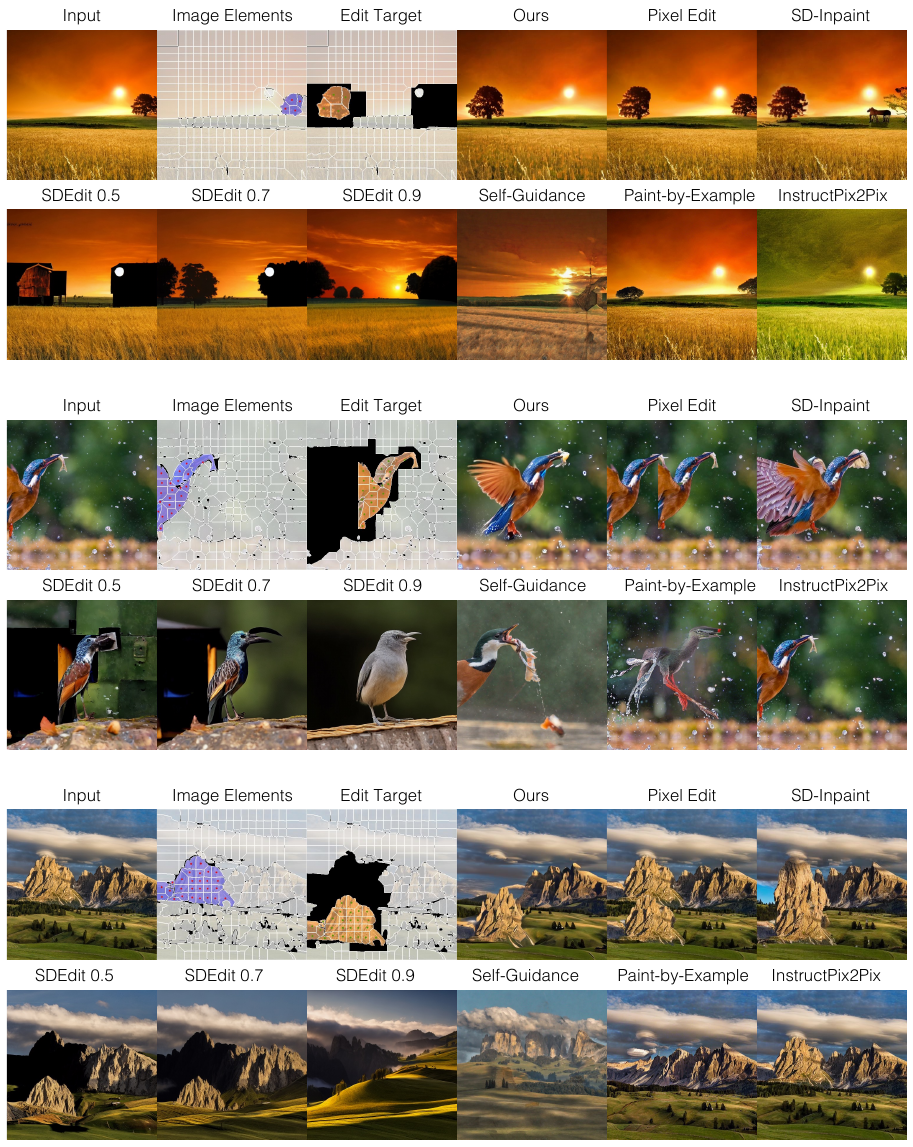}
    \vspace{-0.10in}
    \caption{Comparisons complementing Figure 4, 5, and 6. Our method is compared to pixel editing, pixel editing with Stable Diffusion Inpainting model (SD-Inpaint), pixel editing with SDEdit of various schedules, Self-guidance, Paint-by-Example, and InstructPix2Pix on various edits. Our results attain superior results in preserving the details of the input as well as following the new edits. The baseline results show various types of failures, such as decline in image quality, floating textures, and unfaithful to the edits.}
    \label{supp fig: exp figure 7}
    \vspace{-0.10in}
\end{figure}

\begin{figure}[t]
    \centering
    \includegraphics[width=\linewidth]{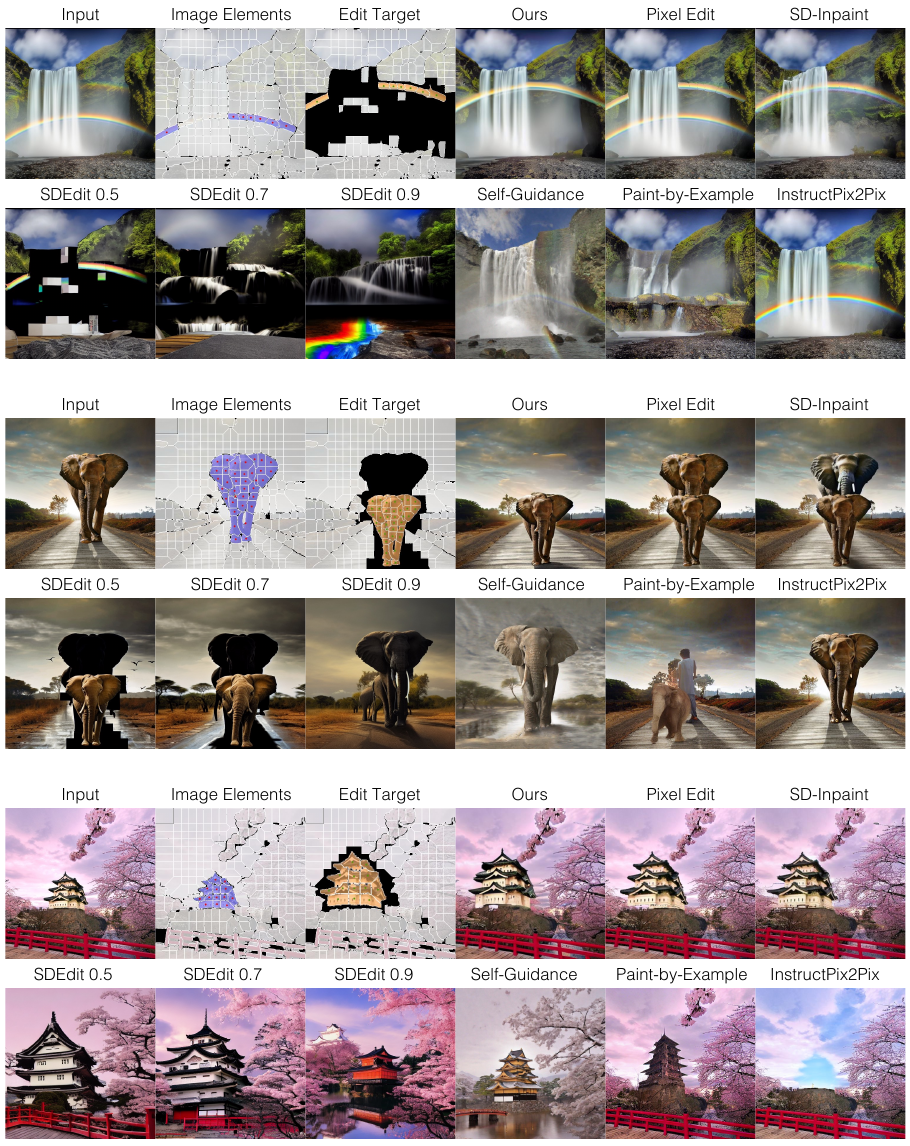}
    \vspace{-0.10in}
    \caption{Comparisons complementing Figure 4, 5, and 6. Our method is compared to pixel editing, pixel editing with Stable Diffusion Inpainting model (SD-Inpaint), pixel editing with SDEdit of various schedules, Self-guidance, Paint-by-Example, and InstructPix2Pix on various edits. Our results attain superior results in preserving the details of the input as well as following the new edits. The baseline results show various types of failures, such as decline in image quality, floating textures, and unfaithful to the edits.}
    \label{supp fig: exp figure 8}
    \vspace{-0.10in}
\end{figure}

%
%

\clearpage

\end{document}